\documentclass[10pt,compsoc]{IEEEtran}
\ifCLASSOPTIONcompsoc
  \usepackage[nocompress]{cite}
\else
  \usepackage{cite}
\fi
\ifCLASSINFOpdf
\else
\fi
\usepackage[hidelinks]{hyperref} %- 1 fehler weniger
\usepackage{tikz,tikz-dimline}
\usepackage{amssymb,amsthm,amsmath}

\pgfplotsset{compat=1.18}

\newtheorem{lemma}{Lemma}
\newtheorem{theorem}{Theorem}
\newtheorem{corollary}{Corollary}

\newcommand{\SOD}{\Omega}
\newcommand{\proofendspace}{\vspace{1mm}}

\usepackage{xparse}

\newcommand{\example}[1]{\vspace{1mm}\noindent{\textbf{Example.}} #1}\vspace{1mm}

\NewDocumentCommand\revised{o+m}{{\textcolor{blue}{\IfValueT{#1}{[#1]} #2}}}
\NewDocumentCommand\additional{o+m}{\textcolor{violet}{\IfValueT{#1}{[#1]} #2}}

\RenewDocumentCommand\revised{o+m}{#2}
\RenewDocumentCommand\additional{o+m}{#2}

\begin{document}
\title{Robustness of Generalized Median Computation for Consensus Learning in Arbitrary Spaces}

\author{Andreas Nienkötter,
        Sandro Vega-Pons,
        Xiaoyi Jiang,~\IEEEmembership{Senior Member,~IEEE}% <-this % stops a space
        \thanks{Andreas Nienkötter is with the Institute for Disaster Management and Reconstruction, Sichuan University-Hongkong Polytechnic University, Chengdu, Sichuan, China.\\
        E-mail: nienkoetter@scu.edu.cn.}
        \thanks{Sandro Vega-Pons is at Lake Worth, FL 33467, USA.\\
        E-mail: sv.pons@gmail.com.}
        \thanks{Xiaoyi Jiang is with the Faculty of Mathematics and Computer Science, University of Münster, Münster, Germany.\\
        E-mail: xjiang@uni-muenster.de.}
        
    }


\markboth{arXiv, 2025}%
{arXiv,2025}



\IEEEtitleabstractindextext{
\begin{abstract}
Robustness in terms of outliers is an important topic and has been formally studied for a variety of problems in machine learning and computer vision. Generalized median computation is \revised[2.3]{a special instance of consensus learning} and a common approach to finding prototypes. Related research can be found in numerous problem domains with a broad range of applications. So far, however, robustness of generalized median has only been studied in a few specific spaces. To our knowledge, there is no robustness characterization in a general setting, i.e. for arbitrary spaces. We address this open issue in our work. The breakdown point $\geq 0.5$ is proved for generalized median with metric distance functions in general. We also study the detailed behavior in case of outliers from different perspectives. In addition, we present robustness results for weighted generalized median computation and non-metric distance functions. Given the importance of robustness, our work contributes to closing a gap in the literature. The presented results have general impact and applicability, e.g. providing deeper understanding of generalized median computation and practical guidance to avoid non-robust computation.
\end{abstract}

}

\maketitle

\IEEEdisplaynontitleabstractindextext

% \fi
\IEEEpeerreviewmaketitle

%\IEEEraisesectionheading{\section{Introduction}\label{sec:introduction}}


%\IEEEraisesectionheading{\section{Introduction}\label{introduction}}

\section{Introduction}
\label{introduction}

%\IEEEPARstart{R}{obustness} 
Robustness has been formally studied for several problems in machine learning \cite{Holland2017,Kanamori2017} and computer vision \cite{Chin2020}.
%Meer_2014,
Here the robustness is meant to be the capacity of tolerating outliers without significant deviation from expected results or even total failure. In this paper we investigate diverse robustness aspects of  generalized median (GM) computation.

\revised[2.3]{Consensus problems have a long history in computer science, appearing in various forms. For example, consensus is a core challenge in cooperative control of multi-agent systems \cite{Rezaee2017}, the group decision-making problem \cite{Meng2017}, and supporting social consensus in discussions on countermeasures for information technology risks \cite{Samejima2015}. In pattern recognition, combining multiple classification methods to reach a consensus decision helps offset the errors of individual classifiers. In practice, ensemble methods have proven to be an effective means of improving classification performance across numerous applications \cite{Yang2023}.}

\revised[2.3]{This paper focuses on another instance of consensus learning, namely learning a prototype from a set of objects, a key problem in machine learning and pattern recognition.}
Given a set of objects $O = \{o_1,...,o_n\}$ in domain $\mathcal{D}$ with an associated distance function $\delta : \mathcal{D} \times \mathcal{D} \rightarrow \mathbb{R}_0^+$, one common approach to finding prototypes (information aggregation) is to determine the so-called \textit{generalized median} \revised[1.22]{\cite{ferrer2010generalized,Jiang2023}}:
\begin{equation}
    \bar{o} = \arg \min_{o \in \mathcal{D}} \sum_{o_i \in O} \delta(o, o_i) \ = \ \arg \min_{o \in \mathcal{D}} \ \SOD_O(o)
\label{eq:median}
\end{equation}
%\begin{align}
%    \bar{o} = \arg \min_{o \in \mathcal{D}} \underbrace{\sum_{o_i \in O} \delta(o, o_i)}_{\SOD_O(o)}
%\label{eq:median}
%\end{align}
%
where $\SOD_O(o)$ is the sum of distance between $o$ and set $O$.
Intuitively, GM represents a formalized averaging. In addition, it can also be understood as a specific form of ensemble solution. In several domains with a specific distance function GM is also known under different names, e.g. 
%geometric median in case of Euclidean vector spaces with the Euclidean distance \cite{ferrer2010generalized} or
Karcher mean in case of positive definite matrices \cite{lim2012matrix} \revised[1.10, 1.22]{or geometric median in Riemannian manifolds \cite{fletcher2009geometric}}. Note that a related concept is medoid (or set median) that results from a similar optimization when restricting the search space to $O$ instead of the complete domain $\mathcal{D}$.

\example{\additional{When dealing with real numbers ($\mathcal{D}=\mathbb{R}$), we have the well-known concepts from statistics. In case of $\delta(p,q) = (p-q)^2, \ p,q \in \mathbb{R}$, the GM is simply the arithmetic average of the given numbers. Changing to another distance function $\delta(p,q) = |p-q|$ results in the median. If restricted to $\mathcal{D}=\mathbb{R}^+$, we obtain the GM: $(x_1 x_2 \cdots x_N)^{1/N}$ ($N$: input size) using the squared hyperbolic distance $\delta(p,q) = |\log p-\log q|$ \cite{Moakher02}.}}

\example{\revised[2.1, 2.6, 2.16]{In the case of 3D rotations, averaging is a common task in computer vision such as structure from motion or calibration \cite{Rotation_Averaging_2013}, where several 3D rotations are measured and combined to get a robust estimation of the target objects true rotation. One common metric for this task is the angular distance \cite{Rotation_Averaging_2013}. Given two rotations $R_1$ and $R_2 \in SO(3)$, it is the smallest rotation angle of the rotation $R_1^T R_2$, i.e. the rotation angle of the rotation that transforms between $R_1$ and $R_2$. This distance can be computed by:
\begin{align*}
    \delta (R_1, R_2) = \lVert \operatorname{Log}(R_1^T R_2) \rVert_F
\end{align*}
where, $\lVert \cdot \rVert_F$ is the Frobenius norm of a matrix and $\operatorname{Log}$ is the matrix logarithm. Often these applications use the mean of 3D rotations, computed with:
\begin{align*}
    \bar{R} = \arg \min_{R \in SO(3)} \sum_{R_i \in O} \delta(R, R_i)^2 
\end{align*}
i.e. the GM with the (non-metric) squared distance $\delta^2$.}}

\example{\revised[2.1, 2.6, 2.16]{Averaging of rankings is a task common in consensus learning, such as combination of classifier rankings, web queries, or biological data \cite{Boulakia_2011}. Here, the GM is used together with the popular Kendall-tau metric
\begin{align*}
    \delta(r_1, r_2) =& \lvert\lbrace (i,j): i < j, \ [r_1(i) < r_1(j) \land r_2(i) > r_2(j)]\\
    &\lor [r_1(i) > r_1(j) \land r_2(i) < r_2(j)]\rbrace\rvert
\end{align*}
to compute a representative ranking. This metric is the number of position pairs $(i,j)$ in rankings $r_1$ and $r_2$ where the ordering differs between them.}}

\vspace{2mm}
GM has been intensively studied for numerous problem domains. Examples include rankings \cite{Boulakia_2011}, phase or orientation data (data smoothing) \cite{Storath_2018,Guo2020}, point-sets \cite{Ding14}, 3D rigid structures \cite{Ding_NIPS2013}, 3D rotations \cite{Rotation_Averaging_2013,Lee2023}, quaternions \cite{Markley2007}, shapes \cite{Cunha_2019}, image segmentation \cite{Khelifi_2017,Ma2021}, clusterings \cite{Carpineto_2012,Boongoen2018},  biclustering \cite{Yin2018}, sequence data (strings) \cite{Mirabal2021}, graphs \cite{jiang2001median,Blumenthal2021}, probability vectors (for ensemble classification) \cite{Lv2022}, anatomical atlas construction \cite{fletcher2009geometric,Xie2013}, averaging in Grassmann manifolds \cite{Chakraborty2021} and \revised[2.4]{flag manifolds \cite{Mankovich_2023}}.

GM computation has found numerous applications in many fields. For instance, consensus ranking is of common interest to bioinformatics, social sciences, and complex network analysis \cite{Posfail2019}. Averaging 3D rotations has been intensively studied for computer vision \cite{Hartley2011}, mixed reality, and computer-assisted surgery \cite{Tu2022}. Such single rotation averaging also serves as a building block for multiple rotation averaging \cite{Lee2022}. Averaging quaternions is interesting for astronautics \cite{Markley2007}.  Ensemble clustering has strong potential for pattern recognition, it is also omnipresent in computer vision \cite{Wazarkar2018}, e.g. in bag-of-words models \cite{Gidaris2020}. Biclustering is fundamental to bioinformatics \cite{Yin2018}. Other applications include robust and nonlinear subspace learning \cite{Hauberg_2014,Chakraborty2021}, robust statistics on Riemannian manifolds \cite{Fletcher2008}, and deep learning (neural network for manifold-valued data \cite{Chakraborty2022}, filter pruning for deep convolutional neural networks acceleration \cite{He2019}), and test time augmentation in deep learning \cite{Bruns2024}.

Several theoretical issues have been studied in the literature. The recent work \cite{Nienkoetter2021} shows a statistical interpretation of GM as a maximum-likelihood estimator. Despite the simple definition in Eq.~(\ref{eq:median}), many instances of GM turn out to be $NP$-hard. A number of papers are concerned with such complexity proofs (e.g. strings \cite{de2000topology},
%rankings with the popular Kendall-$\tau$ distance \cite{Boulakia_2011},
ensemble clustering \cite{Krivanek_1986}, %as well as graphs \cite{ferrer2010generalized}
and signed permutations \cite{bader2011transposition}).
%Even in the very common case of the median vector using the Euclidean distance in $\mathbb{R}^d$ there is no known polynomial-time algorithm. Here, it is not even known if the problem is in $NP$ or even worse \cite{hakimi_2000}.
%\textcolor{red}{Although defined by a simple equation, the computation of the generalized median is $NP$-hard in many cases, for example strings using the string edit distance \cite{de2000topology}. The same applies to median ranking under the generalized Kendall-$\tau$ distance \cite{Boulakia_2011}, ensemble clustering for reasonable clustering distance functions (e.g. a proof is given in \cite{Krivanek_1986} for the Mirkin-metric), median graphs \cite{ferrer2010generalized} and signed permutations \cite{bader2011transposition} using common distance functions, just to name a few examples. Even for the seemingly simple case of $\mathbb{R}^d$ with the Euclidean distance there is no known polynomial-time algorithm, and it is not even known if this problem is in $\mathcal{NP}$ \cite{hakimi_2000}.}
%Nevertheless, the problem of the generalized median has been studied extensively, with solutions ranging from domain specific methods (e.g. clusterings \cite{Carpineto_2012}, graphs \cite{jiang2001median}) to domain independent ones that can be applied to any domain \cite{Franek_SSPR_2012,jiang2001median,nienkoetter2021dpe,ferrer2010generalized}.
Due to the inherent hardness many algorithms for GM computation are of approximate nature so that there is no guarantee to obtain the optimal solution. Since the genuine GM is unknown, a lower bound of the sum of distances $\SOD$ can be used to assess the quality of the computed GM. Several lower bounds are available in general cases \cite{Nienkotter2020} or for specific domains \cite{Goder2008} (consensus clustering).

In this work we study another theoretical issue, namely the robustness of GM computation in general, i.e. without assumption of any particular space. At some places we make the rather weak assumption of metric distance functions. Concretely, we contribute to the following aspects:
\begin{itemize}
\item The breakdown point is a central concept of robust statistics. We show a breakdown point $\geq 0.5$ of GM for arbitrary metric distance functions.
\item While the breakdown point is a global characterization of outlier tolerance, we go a step further and study the detailed behavior in case of outliers for metric distance functions. In particular, we look at the influence of additional and corrupted objects on the location of the GM. Bounds of the maximum displacement of the GM in the case of added and replaced objects will be derived. In addition, we study the sum of distance between original and corrupted median and give an upper bound.
\item \revised[2.5]{In addition to non-weighted GM results, we examine median robustness under the influence of object weights, and give bounds for the breakdown point and maximum displacement of the corrupted median for the weighted GM.}
\item Non-metric distance functions are more difficult to deal with. Still, we will present some robustness results in such spaces. \revised[2.5]{In particular, we show the non-robustness of $\delta^p$, $p$ integer $\geq 2$, for metric distance function $\delta$ (under a weak assumption).}
\end{itemize}
\revised[1.23]{This paper focuses on theoretical results that enhance our understanding of the fundamentals of averaging in arbitrary spaces. Importantly, these results are also highly relevant in practice, given the current lack of awareness regarding the unsuitability of certain distance functions in terms of robustness. As such, our work provides valuable guidelines for practitioners.}

The breakdown point has been studied for some specific spaces, e.g. $\mathbb{R}$, Euclidean spaces \cite{small1990survey} \revised[1.6]{\cite{lopuhaa_breakdown_1991}}, and Riemannian manifolds \cite{fletcher2009geometric}. To our knowledge, however, there is no general consideration for arbitrary spaces so far. The other aspects of our study do not even appear to exist for specific spaces. Given the importance of robustness, our work thus contributes to closing a gap in the literature. Due to the general nature of our work (no assumption of specific space) our results are derived by intensive use of triangle inequality. These results are of theoretical interest and also have general impact and applicability (see the discussion in Section \ref{sec:application}).

The remainder of the paper is structured as follows.
In Section \ref{sec:definition} we introduce the notion of breakdown point used in this paper. 
Section \ref{sec:riemann_proof} proves the breakdown point $\geq 0.5$ of GM for metric distance functions. The detailed behavior study will be given in Section \ref{sec:bound} to \ref{sec:sod_bound}. Then, two extensions follow: weighted GM (Section \ref{sec:weighted-med}) and non-metric distance functions (Section \ref{sec:non-metric}). 
\additional{To enhance readability, we first provide a brief proof sketch and illustrative examples for each theorem in Section \ref{sec:riemann_proof}--\ref{sec:non-metric}. The full proofs of all theorems are collected in Section \ref{sec:proofs} which readers can easily skip if desired.}
The application potential of the theoretical results including two experiments is presented in Section \ref{sec:application}.
\additional{Given the importance of metric distance functions for robust GM computation, we briefly discuss several techniques for obtaining metrics in Section \ref{sec:metric}.}
Finally, some further discussions in Section \ref{sec:conclusion} conclude our paper.

 %%%%%%
\section{Breakdown point}
\label{sec:definition}

Similar to \cite{donoho1983notion,Maronna2019}, the \textit{finite-sample replacement breakdown point} of the GM $\bar{o}$ can be defined as:
\begin{align}
	\epsilon^* = \min_{1 \leq k \leq n} \left\lbrace \frac{k}{n} : \sup_{Q} \delta(\bar{o},\bar{q}) = \infty \right\rbrace \label{eq:breakdown}
\end{align}
where $\bar{q}$ is the GM of a multi-set $Q$ that replaces $k$ objects of $O$ with arbitrary ones. As such, this breakdown point is the minimum ratio of the $k$ corrupted objects to $n = |O|$ original ones that allows the corrupted median $\bar{q}$ to become arbitrary large displaced from the original median $\bar{o}$. In other words, the computation can tolerate at most a fraction $k/n$ of outliers.
\revised[2.14c]{This definition only holds in cases where $\sup_Q \delta(\bar{o}, \bar{c}) = \infty$ exists. Metrics bounded by a constant for example would have no breakdown point and would always be robust towards outliers. In the following, we therefore assume that the breakdown point exists for all further considerations.}

%%%%%%
\section{Breakdown point of GM}
\label{sec:riemann_proof}

\revised[1.6]{In \cite{lopuhaa_breakdown_1991}, the exact breakdown point of $0.5$ of the generalized median was proven for $\mathbb{R}^d$ under the Euclidean distance. This proof was then extended  to Riemannian manifolds in \cite{fletcher2009geometric}, resulting in the same breakdown point of $0.5$.}
In this section we adapt this proof for general metric distances, in particular distances in non-complete and discrete spaces. In this general case, the breakdown point can only be shown to be $\geq 0.5$, \revised[1.10]{i.e. our proof does not rule out that a breakdown point of $>0.5$ might apply to non-trivial special metrics, which would be an interesting outcome.}

\begin{theorem}
Let $\mathcal{D}$ be an arbitrary space with a metric $\delta: \mathcal{D} \times \mathcal{D} \rightarrow \mathbb{R}_0^+$, and $O = \{o_1, ..., o_n\}$ be a multi-set in $\mathcal{D}$. Then, the GM $\bar{o}$ of $O$ has a breakdown point $\epsilon^* \geq \lfloor (n+1)/2 \rfloor/n$ with $\displaystyle\lim_{n\rightarrow \infty} \epsilon^* \geq 0.5$. \additional{An upper bound for the displacement of the median object is $\delta(\bar{o},\bar{q}) \leq (n+1)R + \lfloor(n+1)2\rfloor c$, where $\bar{q}$ is the GM of $O$ with $k \leq \lfloor (n-1)/2 \rfloor$ objects having been replaced by arbitrary outliers, and $c$ is a constant.}
\label{theorem:riemann_bound}
\end{theorem}

\begin{figure}[t]
    \centering
    \begin{tikzpicture}[%
    xscale=1,
    yscale=1,
    setO/.style={circle,minimum size=5pt,inner sep=0mm,draw=black,fill=black},
    setP/.style={circle,minimum size=5pt,inner sep=0mm,draw=black},
    setExample/.style={circle, minimum size=2pt, inner sep=0pt, draw=black, fill=black}]

% X
\node at (0.75,2.6) {$O$};
\node[setO] (o1) at (1,0) {};
\node[setO] (o2) at (0.1,1.8) {};
\node[setO] (o3) at (1.5,2.1) {};
\node[setO] (o4) at (-0.1,0.6) {};

\node[setO,label=left:$\bar{o}$] (xm) at (1,1) {};

\draw[fill=black,opacity=0.1] (-0.3,-0.2) rectangle (1.7,2.8);

\begin{scope}
    \clip(-0.5,-0.4) rectangle (4,3);
    \draw (xm) circle[radius=2.5];
    \draw (xm) circle[radius=1.25];
\end{scope}

%\draw[dashed] (xm) -- node[below left] {$R$} (o2);
\draw[dashed] (xm) -- node[above,pos=0.5] {\footnotesize $R$} +(1.25,0);
\draw[dashed] (xm) -- node[above,pos=0.75] {\footnotesize $R$} +(2.5,0);

% P
\node at (6.7,2.0) {P};
\node[setP] (p1) at (7.2,0.8) {};
\node[setP] (p2) at (6.5,1.5) {};
\node[setP] (p3) at (6.2,0.2) {};

\draw[fill=black,opacity=0.1] (6.0,0.0) rectangle (7.4,2.2);

% Q
\node[setP,label=right:$\bar{q}$] (qm) at (5,1) {};

\node at (4,2.6) {$Q$};
\draw[fill=black,opacity=0.1] (-0.5,-0.4) rectangle (7.6,3.0);

\draw[dashed] (qm) -- node[above] {\footnotesize $\gamma$} +(-1.5,0);

% a inside
\node[setExample, label=below:$a$] (a) at (3.3,0.8) {};
%\draw (qm) -- node[below] {$\gamma$} (a);
\draw[] (a) -- 
    node[below, pos=0.75] {\footnotesize $R$} 
    node[below, pos=0.3] {\footnotesize $<R$} 
    node[setExample, pos=0.47] {} 
    (xm);

% b outside
\node[setExample, label=below:$b$] (b) at (4,0.7) {};
\draw (qm) -- node[below] {\footnotesize $\gamma'$} (b);
\draw (b) -- node[below] {\footnotesize $c$} (a);

% lines
%\draw (o1) -- (om);
%\draw (o2) -- (om);
%\draw (o3) -- (om);
%\draw (o4) -- (om);

%\draw[dashed] (om) -- (qm);

%\draw[dotted] (o1) -- (qm);
%\draw[dotted] (o2) -- (qm);
%\draw[dotted] (o3) -- (qm);
%\draw[dotted] (o4) -- (qm);

%\draw[dashdotted] (p1) -- (qm);
%\draw[dashdotted] (p2) -- (qm);
%\draw[dashdotted] (p3) -- (qm);

\end{tikzpicture}
    \caption{Illustration of corrupted set Q, where objects of the original set $O$ are corrupted (here: moved to $P$). The original median object $\bar{o}$ is displaced to $\bar{q}$. 
    In \cite{fletcher2009geometric} proof, the maximum displacement $\delta(\bar{o},\bar{q})$ is bounded by $2R + \gamma$ (dashed), where $R$ is the maximum distance of any object in $O$ to $\bar{o}$. In discrete spaces we instead show that $\delta(\bar{o}, \bar{q})$ is bounded by $2R + c + \gamma'$ (solid), where $c$ is realized by two objects $a$ and $b$, as an object on the radius $2R$ may not exist.}
    \label{fig:inequality_riemann}
\end{figure}
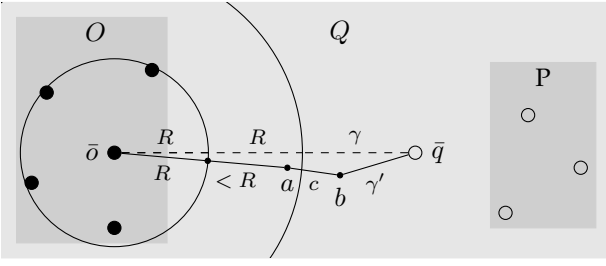

\noindent\revised[1.17]{\emph{Proof sketch.} The proof generally follows \cite{fletcher2009geometric}, with adaptions for general spaces. Given an original set $O$ and a corrupted set $Q$, where $k<\lfloor (n-1)/2 \rfloor$ objects of $O$ have been moved to $P$ (see Figure \ref{fig:inequality_riemann}), \cite{fletcher2009geometric} shows that in Riemannian spaces $\delta(\bar{p}, \bar{q})$ is bounded by $2R + \gamma$ with $\gamma < \infty$, where $R$ is the maximum distance between $\bar{o}$ and any $o_i \in O$. 
This requires an object on the imagined line between $\bar{o}$ and $\bar{q}$ with $\delta(\bar{o}, a) \leq 2R$ and $\delta(a, \bar{q}) = \gamma$. 
In discrete spaces this point may not be readily available. Instead, we show that through two objects $a$ and $b$, which are on the inside and outside of the ball $B$, the inequality $\delta(\bar{o}, \bar{q}) < 2R + c + \gamma'$ can be realized instead with $c = \delta(a,b) < \infty$ and $\gamma' < \infty$.}

\example{\additional{Given the set of four rankings $O = \{[1, 2, 4, 3, 5], [1, 2, 3, 5, 4], [2, 1, 3, 4, 5], [1, 3, 2, 4, 5]\}$, a generalized median using the Kendall-tau distance is $\bar{o} = [1, 2, 3, 4, 5]$ with distance 1 to each element in the set. Therefore, the radius $R$ is $R = 1$. Using Theorem \ref{theorem:riemann_bound}, we can see that replacing an outlier ranking to the set, such as $[5,4,3,2,1]$, would result in a maximum displacement of $\delta(\bar{o}, \bar{q}) \leq (n+1)R+\lfloor(n+1)2\rfloor c = 15$. In this case, $c=1$ since the minimum distance between any two objects is 1. The displaced median therefore is at most 15 disagreements displaced from the original median.}}

With Theorem \ref{theorem:riemann_bound}, it is especially true that for the popular Minkowski distances \revised[1.11]{$\delta(x,y)$} = $\left(\sum_{i=1}^m | x_i - y_i |^ p\right)^{1/p}$, which are metric over $\mathbb{R}^m$, the breakdown point is $\epsilon^* \geq 0.5$. Considering the special case $p=1$, this result directly confirms the breakdown point 0.5 for the common definition of the median in $\mathbb{R}$. The applicability of Theorem \ref{theorem:riemann_bound} is far more than this since (meaningful) metric distance functions exist in many spaces. For example, the string space with the popular string edit distance \cite{Wagner1974} and the graph space with similar graph edit distance \cite{Gao2010} satisfy the metric requirement as well (under very weak assumptions about the cost of the edit operations). Moreover, given a positive definite kernel $K(x,y)$ in any space, one can compute a metric distance $\delta_K(x,y)$ in a Hilbert space $\mathcal{H}$ \cite{Gartner2004}.

%%%%%%
\section{Influence of additional and corrupted objects on location of GM}
\label{sec:bound}

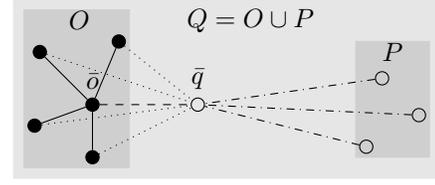
\begin{figure}
    \centering
    \begin{tikzpicture}[%
    xscale=0.7,
    yscale=0.7,
    setO/.style={circle,minimum size=5pt,inner sep=0mm,draw=black,fill=black},
    setP/.style={circle,minimum size=5pt,inner sep=0mm,draw=black}]

% O
\node at (0.75,2.6) {$O$};
\node[setO] (o1) at (1,0) {};
\node[setO] (o2) at (0,2) {};
\node[setO] (o3) at (1.5,2.2) {};
\node[setO] (o4) at (-0.1,0.6) {};

\node[setO,label=$\bar{o}$] (om) at (1,1) {};

\draw[fill=black,opacity=0.1] (-0.3,-0.2) rectangle (1.7,2.8);

% P
\node at (6.7,2.0) {$P$};
\node[setP] (p1) at (7.2,0.8) {};
\node[setP] (p2) at (6.5,1.5) {};
\node[setP] (p3) at (6.2,0.2) {};

\draw[fill=black,opacity=0.1] (6.0,0.0) rectangle (7.4,2.2);

% Q
\node[setP,label=$\bar{q}$] (qm) at (3,1) {};

\node at (4,2.6) {$Q = O \cup P$};
\draw[fill=black,opacity=0.1] (-0.5,-0.4) rectangle (7.6,3.0);

% lines
\draw (o1) -- (om);
\draw (o2) -- (om);
\draw (o3) -- (om);
\draw (o4) -- (om);

\draw[dashed] (om) -- (qm);

\draw[dotted] (o1) -- (qm);
\draw[dotted] (o2) -- (qm);
\draw[dotted] (o3) -- (qm);
\draw[dotted] (o4) -- (qm);

\draw[dashdotted] (p1) -- (qm);
\draw[dashdotted] (p2) -- (qm);
\draw[dashdotted] (p3) -- (qm);

\end{tikzpicture}
    \caption{Illustration of the original set $O$ and added objects $P$. The original median object $\bar{o}$ is displaced to $\bar{q}$ by the inclusion of the added objects in $P$. Note that $\bar{o}$ and $\bar{q}$ are not part of the sets but median objects, minimizing $\SOD_O$ and $\SOD_Q$, respectively.}
    \label{fig:inequality_added}
\end{figure}

In this section we will show a bound of the maximum displacement of the GM in case of added and replaced objects. In the first case (Theorem \ref{theorem:added_bound}) the number of objects of the original set are increased by additional outliers. In the second case (Theorem \ref{theorem:replaced_bound}) the original objects are replaced, thereby keeping the total number of objects unchanged.

\begin{theorem}
Let $\mathcal{D}$ be an arbitrary space with a metric $\delta: \mathcal{D} \times \mathcal{D} \rightarrow \mathbb{R}_0^+$, $O = \{o_1, ..., o_n\}$ a multi-set in $\mathcal{D}$ with GM $\bar{o}$, $P = \{p_1, ..., p_k\}$ a multi-set in $\mathcal{D}$ with $k < n$ and $Q = O \cup P$ with GM $\bar{q}$. Then, $\delta(\bar{o}, \bar{q}) \leq \frac{2}{n-k} \SOD_O(\bar{o})$ holds.
\label{theorem:added_bound}
\end{theorem}

\noindent
\additional{\emph{Proof sketch.} See Figure \ref{fig:inequality_added} for illustration. The proof is entirely depending on the triangle inequality, especially that $\delta(\bar{q},\bar{o}) \leq \delta(\bar{q},o_i) + \delta(o_i, \bar{o})$ for all $o_i$ and $\delta(\bar{q},p_i) - \delta(p_i, \bar{o}) \leq \delta(\bar{q},\bar{o})$ for all corrupted objects $p_i$. Combining these individual bounds of the displacement for $O$ and $P$ bounds the displacement of $Q = O \cup P$ by the sum of distances of $O$.}
\proofendspace

Note that the maximum displacement in this theorem is bounded by the sum of distance of $O$ alone, independent of the location of the objects in $P$.

\example{\additional{Given the same set of rankings as in the example for Theorem \ref{theorem:riemann_bound}, $O = \{[1, 2, 4, 3, 5], [1, 2, 3, 5, 4], [2, 1, 3, 4, 5], [1, 3, 2, 4, 5]\}$, a generalized median using the Kendall-tau distance is $\bar{o} = [1, 2, 3, 4, 5]$ with distance 1 to each element in the set. Therefore, the sum of distance is $\Omega_O(\bar{o}) = 4$. Using Theorem \ref{theorem:added_bound}, we can see that \emph{adding} an outlier ranking to the set, such as $[5,4,3,2,1]$, would result in a maximum displacement of $\delta(\bar{o}, \bar{q}) \leq \frac{2}{n-k} \Omega_O(\bar{o}) = \frac{2}{4-1} 4 = 2\frac{2}{3}$. The displaced median therefore is at most 2 disagreements displaced from the original median, as this distance only has integer values.}}

\begin{theorem}
Let $\mathcal{D}$ be an arbitrary space with a metric $\delta: \mathcal{D} \times \mathcal{D} \rightarrow \mathbb{R}_0^+$, $O = \{o_1, ..., o_n\}$ a multi-set in $\mathcal{D}$ with GM $\bar{o}$, and $Q$ a multi-set in $\mathcal{D}$ with GM $\bar{q}$, where \revised[2.9]{$k \leq \lfloor (n-1)/2 \rfloor$ objects} of $O$ were replaced with arbitrary objects of $\mathcal{D}$. Let $X = O \bigcap Q$ be the objects in $O$ that were not replaced. Then, $\delta(\bar{o}, \bar{q}) \leq \frac{4}{n-2k} \SOD_X(\bar{x})$ holds \revised[2.9]{where $\bar{x}$ is the GM of $X$}.
\label{theorem:replaced_bound}
\end{theorem}

\noindent
\additional{\emph{Proof sketch.} This proof follows by using the triangle inequality $\delta(\bar{o}, \bar{q}) \leq \delta(\bar{x}, \bar{o}) + \delta(\bar{x}, \bar{q})$, and both terms being bounded by Theorem \ref{theorem:added_bound}.}
\proofendspace

Note that this bound relies on the much smaller sum of distance of the uncorrupted part of the set instead of the full $O$ or $Q$. Because $n-k>k$, the bound
breaks down at $k \geq \lfloor(n+1)/2\rfloor$. Therefore, the breakdown point is $\epsilon^* \geq \lfloor(n+1)/2\rfloor/n$ or $\geq 0.5$ similar to Section \ref{sec:riemann_proof}. In total we thus have two different proofs of breakdown point $\geq 0.5$.

\example{\additional{Following the same example as above, \emph{replacing} one of the four rankings with the outlier [5,4,3,2,1] would lead to the bound $\delta(\bar{o}, \bar{q}) \leq \frac{4}{n-2k} \Omega_X(\bar{x}) = \frac{4}{4-2} 3 = 6$, i.e the new generalized median would be displaced by at most 6 disagreements. This is a large improvement over the bound in Theorem \ref{theorem:riemann_bound} which resulted in 15.}}

%%%%%%
\section{Comparison of the upper bounds of displacement for the GM}
\label{sec:median_bound_comparison}

Although both Section \ref{sec:riemann_proof} and \ref{sec:bound} reach the same breakdown point for the GM, they lead to different upper bounds of the displacement of the GM:
\begin{align*}
    \delta(\bar{o},\bar{q}) &\leq (n+1)R + \lfloor(n+1)2\rfloor c & \text{(Eq. (\ref{eq:riemann_bound}))}\\
    \delta(\bar{o},\bar{q}) &\leq \frac{4}{n-2k} \SOD_X(\bar{x}) & \text{(Theorem \ref{theorem:replaced_bound})}
\end{align*}
where $R$ is the maximum distance of $\bar{o}$ to any object in the uncorrupted object set $O$, $c>0$ a constant, and $X$ is the uncorrupted part of the object set.

% removed other bound
%Another bound for the maximum displacement can be found in Theorem 28.2 in \cite{Hsu_2016} based on the results in \cite{Minsker_2015}. By substituting $w_*$ in \cite{Hsu_2016} with $\bar{o}$ and $\bar{y}$ with $\bar{q}$, one can reach the bound
%%
%\begin{align*}
%    \delta(\bar{o}, \bar{q}) &\leq \left( 1 + \frac{1}{2 \alpha} \right) \Delta_Q (\bar{o}, \alpha) & \text{\cite{Hsu_2016}}
%\end{align*}
%%
%where $\Delta_Q (\bar{o}, \alpha)$ is the minimum radius of a ball around $\bar{o}$ such that a fraction of $0.5+\alpha,~ \alpha \in (0,\frac{1}{2})$, objects of $Q$ are inside the ball.

In practice, the bound shown in Theorem \ref{theorem:replaced_bound} is much smaller than Eq.~(\ref{eq:riemann_bound}) even if $c=0$ due to the fact that Theorem \ref{theorem:replaced_bound} only includes the sum of distance of the \textit{uncorrupted} objects and is reduced by the fraction the more original objects are left, while Eq.~(\ref{eq:riemann_bound}) includes $n+1$ times the \textit{maximum} distance of the original objects without being influenced by the actual ratio of corrupted objects. 
%The bound shown in \cite{Hsu_2016} in practice also leads to a worse result, especially in the case of a large number of outliers. Due to the fact that a fraction of $0.5 + \alpha$ objects have to be included in the ball around $\bar{o}$, for a large number of outliers (approaching $k=\frac{n}{2}$) the ball will need a radius approaching infinity as the outliers approach infinity. If outliers are not to be included in the ball, $\alpha$ has to approach 0 causing $\frac{1}{2\alpha}$ to approach infinity instead.

\begin{figure}
    \centering
    \hspace*{-3mm}
    \includegraphics[width=0.5\linewidth]{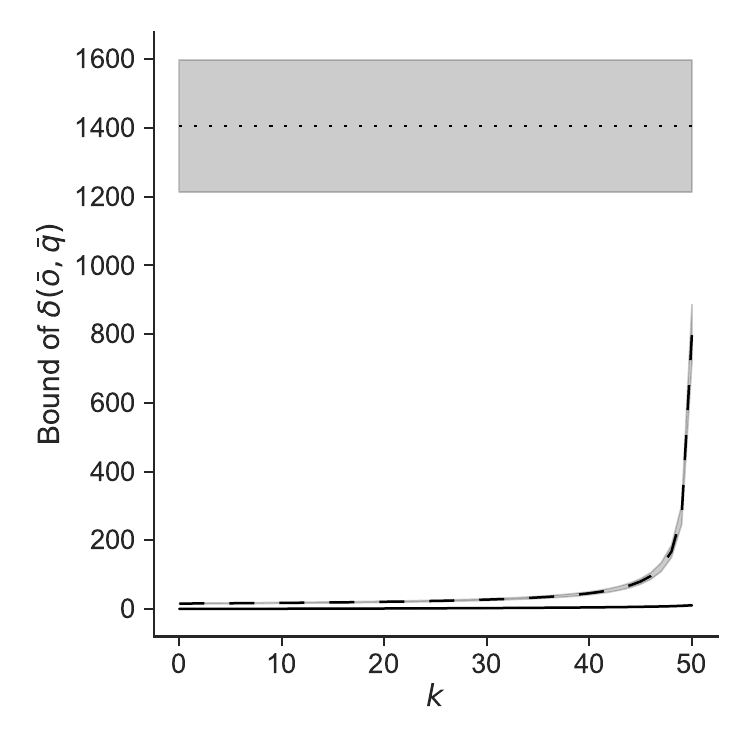}
    \includegraphics[width=0.5\linewidth]{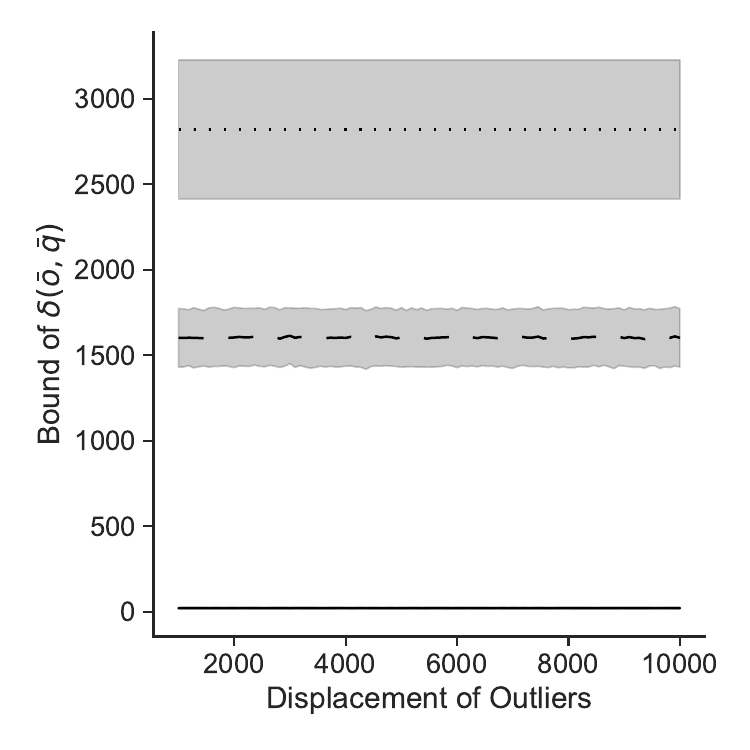}
    \vspace{-1em}
    \caption{\revised[2.10]{$\delta(\bar{o}, \bar{q})$ (solid) compared with the upper bounds resulting from Theorem \ref{theorem:replaced_bound} (ours, dashed) and Eq.~(\ref{eq:riemann_bound}) (dotted)
    for normally distributed datasets, depending on the number (left) and displacement (right) of outliers. In left the displacement is 1000, in right $k=50$.}}
    \label{fig:bounds_outliers}
\end{figure}

\revised[2.10]{The difference in bounds can be seen in Figure \ref{fig:bounds_outliers}. In this example the maximum bounds for a normally distributed dataset of 101 objects from $\mathbb{R}$ with standard deviation 5 are shown. The distance function is set to $\delta(x,y)=|x-y|$. Since there are 101 original objects, the amount of outliers $k$ must be in the range $[0,50]$ for all bounds to hold. 
%For the bound derived from \cite{Hsu_2016}, $\alpha$ was chosen to minimize the bound. 
The shown results are averages and standard deviations of 1000 trials. One can clearly see the much tighter bound resulting from Theorem \ref{theorem:replaced_bound}, especially in cases of low $k$. Unsurprisingly, both bounds are independent of the distance of outliers.
%The only case where this bound is beaten is in the case of a very small displacement of objects where the impact of outliers is small in general. Overall, the bound \cite{Hsu_2016} grows linearly for $k=n/2$ in the displacement of objects, as an outlier has to be included in the bound computation.
}

%%%%%%
\section{Example of a set reaching the bound in Theorem \ref{theorem:added_bound}}
\label{sec:example-set}

In this section we will present a configuration of objects in $\mathbb{R}$ using the Euclidean distance, for which the bound shown in Theorem \ref{theorem:added_bound} is maximized, thereby showing that it is a tight upper bound. 

Consider the two multi-sets:
\begin{itemize}
	\item  $O \subset \mathbb{R}$ with
	$|O| = n_1 + n_2,\;\; n_1 > n_2,$ \\
	$o_i = \begin{cases}
	0, & i = 1,...,n_1 \\
	d \in \mathbb{R}, & i = n_1 + 1, ..., n_1+n_2
	\end{cases}$
	\item $P \subset \mathbb{R}$ with
	$|P| = k,\;\; p_i >> d,\;\; i = 1,...,k$
\end{itemize}
using $n_1,n_2,k > 0$ and distance function $\delta(o_i,o_j) = |o_i - o_j|$.
\revised[2.11]{For $k < n_1+n_2$,} the GMs and sum of distances according to the bound shown in Section \ref{sec:bound} are:
\[
\bar{o} = 0, \
\SOD_O(\bar{o}) = n_2 \cdot d, \
\bar{q} = \begin{cases}
0, & \text{if } n_1 > n_2 + k \\
d, & \text{if } n_1 < n_2 + k \\
\epsilon \in [0,d] & \text{if } n_1 = n_2 + k
\end{cases}
\]
\revised[2.11]{This is because $n_1$ is the amount of summands that are $0$ in the sum of distance to $O$, $n_2$ is the amount that are $d$, and $k$ being the amount of values $>> d$ through the values in $P$. In the first case, there are more $0$ than $\geq d$, leading to the median being $0$. In the second case, there are more $\geq d$ than $0$, but $0$ and $d$ still providing the majority of values, leading to the median $d$. In the third case, there is an equal number of $\geq d$ and $0$, with $0$ and $d$ providing the majority, leading to any value in the interval between them being a median.}
In the case of $n_1 > n_2 > 1$ and $k = n_1 - n_2 + 1$, the GM of $Q = O \cup P$ is $\bar{q} = d$. And it follows using Eq.~(\ref{eq:7}):
\begin{align}
& d \ = \ \delta(\bar{o}, \bar{q}) \ \leq \ \frac{2}{n-k} \SOD_O(\bar{o}) \ = \ \frac{2}{n-k} n_2 \cdot d \nonumber \\
& = \frac{2 n_2 d}{n_1+n_2-k} \ = \ \frac{2 n_2 d}{n_1 + n_2 - (n_1 - n_2 + 1)} \nonumber \\
& = \frac{2 n_2}{2 n_2 - 1} d \label{eq:example_bound}
\end{align}
$\frac{2 n_2}{2 n_2 - 1}$ is monotonously falling for $n_2 > 0.5$ and approaches $1$ for $n_2 \rightarrow \infty$. Therefore,
\begin{align*}
 \lim\limits_{n_2 \rightarrow \infty} \frac{2}{n-k} \SOD_O(\bar{o}) = \delta(\bar{o}, \bar{q})
\end{align*}
and the boundary can be reached arbitrarily close.

\begin{figure}
    \centering
    \begin{tikzpicture}[scale=0.6]

\begin{axis}[%
    width=9cm,
    height=5cm,
    xlabel=$n_2$,
    ylabel={Upper bound},
    ytick={1,2},
    yticklabels={$d$,$2d$}
    ]

\addplot [
    domain=1:100, 
    color=black
    ]
    {(2*x)/(2*x-1)};
\end{axis}

\end{tikzpicture}
    \vspace{-1em}
    \caption{The upper bound for $\delta(\bar{o},\bar{q})$ in relation to $n_2$ for the example set shown in Section \ref{sec:example-set}. \revised[2.11]{As shown in (\ref{eq:example_bound}), this relation holds for all $n_1, k$ with $n_1 > n_2$ and $k = n_1 - n_2 + 1$.}}
    \label{fig:example}
\end{figure}
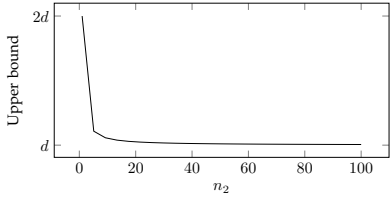

This can be seen in Figure \ref{fig:example}. Although the bound of $\frac{2}{n-k} \SOD_O(\bar{o})$ can be large for arbitrary sets, in this example it quickly converges to the true value $d$ with increased number of objects.

%%%%%%
\section{Upper bound of the sum of distance between original and corrupted median}
\label{sec:sod_bound}

In this section we will show an upper bound for the difference in sum of distances between the true and corrupted GM in case of added objects.

\begin{theorem}
Let $\mathcal{D}$ be an arbitrary space with a metric $\delta: \mathcal{D} \times \mathcal{D} \rightarrow \mathbb{R}_0^+$, $O = \{o_1, ..., o_n\}$ a multi-set in $\mathcal{D}$ with GM $\bar{o}$, $P = \{p_1, ..., p_k\}$ a multi-set in $\mathcal{D}$ with $k < n$, and $Q = O \cup P$ with GM $\bar{q}$. Then, $\SOD_Q(\bar{o}) - \SOD_Q(\bar{q}) < \frac{2 k}{n-k} \SOD_O(\bar{o})$ holds.
\label{theorem:sod_bound}
\end{theorem}

\noindent\additional{\emph{Proof sketch.} The proof is a direct application of the triangle inequality and Theorem \ref{theorem:added_bound}.}

\example{\additional{Given the same set of rankings as in the previous examples, $O = \{[1, 2, 4, 3, 5], [1, 2, 3, 5, 4], [2, 1, 3, 4, 5], [1, 3, 2, 4, 5]\}$, a generalized median using the Kendall-tau distance is $\bar{o} = [1, 2, 3, 4, 5]$. Given a set $Q = O \cup P$ with $P=\{[5,4,3,2,1], [5,4,3,2,1], [5,4,3,2,1]\}$, a generalized median of $Q$ is $\bar{q} = [1,3,2,5,4]$. Then, Theorem \ref{theorem:sod_bound} shows that the difference in $\Omega_Q(\bar{o})$ and $\Omega_Q(\bar{p})$ is smaller than $\frac{2k}{n-k}\Omega_O(\bar{o}) = 24$. Therefore, the sum of distance from the original median to the corrupted median of the corrupted set can at most increase by 24. In our case, it is $\Omega_Q(\bar{o}) = 34$ and $\Omega_Q(\bar{q})=32$ with a difference of 2.}}

This provides an alternative view of the difference in the original and new median object.

%%%%%%

\section{Robustness of weighted GM}
\label{sec:weighted-med}

In the case of weighted GM (e.g. \cite{Fletcher2008,Lv2022,Zhang2022}),
\[
    \bar{o} \ = \ \arg \min_{o \in \mathcal{D}} \sum_{o_i \in O} w_{o_i} \delta(o, o_i) \ = \ \arg \min_{o \in \mathcal{D}}  \Omega_O^w(o)
\]
with weights $w_{o_i} > 0$ for each $o_i$, the robustness cannot be guaranteed in the general case. This can be shown by the following counterexample. Given three numbers $o_1 = 1, o_2 = 1$ and $o_3=2$ with weights $w_1 = 1, w_2 = 1, w_3 = 3$, the weighted GM using distance $\delta(o_i, o_j) = \lvert o_i - o_j \rvert$ is $\bar{o} = o_3 = 2$. By corrupting just $o_3$ to move arbitrarily far away, the new median $\bar{q}$ will follow $o_3$ to have an arbitrarily large distance from the original $\bar{o}$. \revised[2.13]{This is due to $o_3$ having a weight greater than $o_1$ and $o_2$ combined that ensures that any deviation of the median from $o_3$ is increasing the weighted sum of distances.} Therefore, only one object has to be changed in this case to cause the weighted median to break down. This proof can be easily extended to arbitrary $n$ by duplicating $o_1$ and increasing the weight of $o_3$ accordingly.

This breakdown is caused by the fact that the weight of the corrupted object has a greater influence than the weights of the uncorrupted objects, thereby having a majority of the influence on the sum of distances. In the following theorem we will show that the weighted GM has a breakdown point relative to the weights of the uncorrupted objects.

\begin{theorem}
\revised[2.13a]{Let $\mathcal{D}$ be an arbitrary space with a metric $\delta: \mathcal{D} \times \mathcal{D} \rightarrow \mathbb{R}_0^+$, $O = \{o_1, ..., o_n\}$ a multi-set in $\mathcal{D}$ with weights $w_{o_i}$ and weighted GM $\bar{o}$, $P = \{p_1, ..., p_k\}$ a multi-set in $\mathcal{D}$ with weights $w_{p_i}$, $k < n$, and $Q = O \cup P$ with weighted GM $\bar{q}$. Then, $\delta(\bar{o}, \bar{q}) \leq \frac{2}{\sum_{o_i \in O} w_{o_i} - \sum_{p_i \in P} w_{p_i}} \SOD_O(\bar{o})$ holds if $\sum_{o_i \in O} w_{o_i} > \sum_{p_i \in P} w_{p_i}$}.
\label{theorem:added_bound_weighted}
\end{theorem}

\noindent
\additional{\emph{Proof sketch.} This proof follows the same idea as Theorem \ref{theorem:added_bound}. Due to being the weighted GM, the numbers $k$ and $n$ of original and added objects are ultimately replaced by the sum of their respective weights.}
\proofendspace

\example{\additional{Given the set of four rankings $O = \{[1, 2, 4, 3, 5], [1, 2, 3, 5, 4], [2, 1, 3, 4, 5], [1, 3, 2, 4, 5]\}$ from the example for Theorem \ref{theorem:added_bound} with weights $w = \{1,2,1,1\}$, a weighted generalized median using the Kendall-tau distance is again $\bar{o} = [1, 2, 3, 4, 5]$ with weighted distance 1 to each element in the set, and weighted distance 2 to the second ranking. Therefore, the sum of weighted distance is $\Omega_O(\bar{o}) = 5$. Using Theorem \ref{theorem:added_bound_weighted}, we can see that by \emph{adding} an outlier ranking to the set, such as $[5,4,3,2,1]$ with weight $w$, would result in a maximum displacement of $\delta(\bar{o}, \bar{q}) \leq \frac{2}{5-w} \Omega_O(\bar{o}) = \frac{2}{5-w} 5 = \frac{10}{5-w}$. As such, the median is robust if the weight of the outlier is $w < 5$ and the maximum displacement would be $\frac{10}{5-w}$ depending on the outlier. In the case of $w=2$, for example, the displacement would be at most $3 \frac{1}{3}$ or $3$ since only integer distances are possible.}}

\begin{theorem}
\revised[2.13a]{Let $\mathcal{D}$ be an arbitrary space with a metric $\delta: \mathcal{D} \times \mathcal{D} \rightarrow \mathbb{R}_0^+$, $O = \{o_1, ..., o_n\}$ a multi-set in $\mathcal{D}$ with weights $w_{o_i}$ and weighted GM $\bar{o}$, and $Q$ a multi-set in $\mathcal{D}$ with weighted GM $\bar{q}$, where $k$ objects of $O$ were replaced with arbitrary objects of $\mathcal{D}$. Let $X = O \bigcap Q$ be the objects in $O$ that were not replaced, and $P = Q \setminus X$ the corrupted objects. Then, $\delta(\bar{o}, \bar{q}) \leq \frac{4}{\sum_{x_i \in X} w_{x_i} - \sum_{p_i \in P} w_{p_i}} \SOD^W_X(\bar{x})$ holds, where $\bar{x}$ is the weighted GM of $X$, if $\sum_{p_i \in P} w_{p_i} < \sum_{x_i \in X} w_{x_i}$.}
\label{theorem:replaced_bound_weighted}
\end{theorem}

\noindent\additional{\emph{Proof sketch.} This proof follows the same idea as Theorem \ref{theorem:replaced_bound}, using the weighted bound of Theorem \ref{theorem:added_bound_weighted}.}
\proofendspace

\example{\additional{Following the example above, \emph{replacing} a ranking with an outlier would lead to the following cases. If one of the rankings with weight 1 is replaced, then the bound would be $\delta(\bar{o}, \bar{q}) \leq \frac{4}{4 - 1} 4 = \frac{16}{3} = 5 \frac{1}{3}$, i.e. a displacement of at most 5 disagreements. If, however, the ranking with weight 2 is replaced, then the result would be $\delta(\bar{o}, \bar{q}) \leq \frac{4}{3 - 2} 3 = 12$, i.e. at most $12$ disagreements. In both cases, the median is robust as the weight of the outlier is smaller than the combined weights of the remaining set.}}

\revised[2.14b,c]{%
These results show that under certain conditions the weighted GM is not only robust, but the maximum number of objects allowed to be added or replaced until these bounds no longer hold depends only on the weights of $O$ and $P$. Since it is generally unknown which of the objects are the outliers and which are not, one can estimate the breakdown point of the weighted GM for a specific set of weights $W$ by finding the maximum $k=|P|$ so that:
\begin{align*}
    \sum_{p_i \in P} w_{p_i} < \sum_{o_i \in O} w_{o_i}
\end{align*}
holds for any division of $P$ and $O$. This is easy to compute by assigning weights to $P$ by order of size. The breakdown point for this set of weights $W$ is then $\epsilon^* \geq k/n$.}

Note that violating the above assumption $\sum_{p_i \in P} w_{p_i} < \sum_{x_i \in X} w_{x_i}$ is no guarantee that the weighted GM is not robust. If the metric function:
\begin{align*}
    \delta(x,y) = \begin{cases}
    0,& \text{if }x = y \\
    1,& \text{otherwise} \\
    \end{cases}
\end{align*}
is used then $\delta(\bar{o}, \bar{q}) \leq 1$ regardless of the fraction of corrupted objects and weights of the involved objects. Using this metric the weighted GM is always robust \revised[2.14d]{and has no breakdown point}.

%%%%%%
\section{Results for non-metric distance functions}
\label{sec:non-metric}

All the results shown so far are related to metric distance functions. In this section we show some results for non-metric distance functions. Here the results differ in different settings. The GM with a non-metric distance function can be robust. But for a particular class of non-metric distance function the breakdown point is shown to be zero.

\begin{lemma}
The GM with a non-metric distance function can be robust.
\label{lemma:1}
\end{lemma}

\noindent
\revised[1.12]{\textbf{Example.}}
This result can be easily demonstrated with an example in the domain of integers. Given some small constant integer c, the set of all integers is divided into two subsets $S_1 = \{ -c, -c+1, \ldots, c-1, c\}$ and $S_2 = \{\mbox{all other integers}\}$. We define the distance function as:
\[
\delta(x,y) =
\left\{
\begin{array}{ll}
(x-y)^2 & x, y \in S_1 \\
|x-y| & \mbox{otherwise}
\end{array}
\right.
\]
This distance function is non-metric due to the violation of the triangle inequality by the quadratic distance. Since $S_1$ is a small set only, the breakdown point is fully determined by the infinite set $S_2$ and the related metric absolute distance. Thus, in this case the GM is robust, \revised [2.15a]{i.e. with breakdown point $\epsilon^* \geq 0.5$}.

\revised[1.15]{This simple example can be easily extended to a more general case with $\mathcal{D}=\mathbb{R}^d$. The distance function above is accordingly extended to:
\[
\delta(x,y) =
\left\{
\begin{array}{ll}
||x-y||^2 & x, y \in HS_c \\
||x-y|| & \mbox{otherwise}
\end{array}
\right.
\]
where $HS_c$ represents the $d$-D hypersphere with radius $c$. Similarly, the GM is robust in this case.
}

\vspace{1mm}
For the following discussion we need the concept of weighted mean. Given two objects $x$ and $z$, a weighted mean $y$ is intuitively an object between them and is formally defined with weight $0 \leq w \leq 1$ by:
\begin{equation}
\delta(x,y) \ = \ \delta(x,z) + \delta(z,y)
\label{eq:wm}
\end{equation}
\[
\delta(x,y) = w \cdot \delta(x,z), \quad \delta(y, z) = (1-w) \cdot \delta(x,z)
\]
\revised[1.11]{\example{We consider the vector space $\mathcal{D}=\mathbb{R}^d$ together with the Euclidean distance $\delta(p,q) = ||p-q||$. The weighted mean is simply $(1-w)x+wy$ for all $x,y \ (x \not= y)$.
The weighted mean also exists in discrete spaces, such as strings \cite{Bunke2002}, graphs \cite{Bunke2001}, clusterings \cite{Franek2014}, and graph correspondences \cite{Moreno2020}. The related metric distance function is string edit distance \cite{Wagner1974}, graph edit distance \cite{Bunke1983}, partition distance for clusterings \cite{Gusfield2002}, and correspondence edit distance, respectively \cite{Moreno2020}.
}}

\revised[1.11]{
In case of a metric distance function the triangle inequality holds:
\[
\delta(x,y) \leq \delta(x,z) + \delta(z,y)
\]
The existence of a weighted mean (\ref{eq:wm}) represents the weakest condition for satisfying the triangle inequality. Modifying the distance function may, therefore, disrupt this property. In Lemma \ref{lemma:2}, we demonstrate that the powered distance function is one such modification.
}

\begin{lemma}
We consider a space $\mathcal{D}$ with an associated metric distance $\delta$. It is assumed that there exists at least one triple of distinct objects $x$, $y$, and $z$ such that $y$ is a weighted mean of $x$ and $z$, i.e. $\delta(x,y) + \delta(y,z) = \delta(x,z)$. \revised[1.12]{Under this condition, $\delta^p$, where $p$ is an integer with $p \ge 2$, is no longer a metric}.
\label{lemma:2}
\end{lemma}

\noindent
It is easy to show the violation of the triangle inequality.
\begin{eqnarray*}
\delta^p(x,y) + \delta^p(y,z) & = & (\delta(x,y) + \delta(y,z))^p \\
& & - \sum_{k=1}^{p-1} C_p^k \delta^k(x,y) \delta^{p-k}(y,z) \\
& < & (\delta(x,y) + \delta(y,z))^p \\
& = & \delta^p(x,z)
\end{eqnarray*}
Note that although violating the triangle inequality, the non-metric $\delta^p$ still satisfies all other properties of metrics.

{\example{\additional{
%We consider real numbers ($\mathcal{D}=\mathbb{R}$). For the metric distance function $\delta(p,q) = |p-q|$, the weighted mean $(1-w)x+wy$ exists for all $x,y \ (x \not= y)$ (which is stronger than required by Lemma \ref{lemma:2}). In this case the squared distance function $\delta^2(p,q) = (p-q)^2$ is no longer a metric.
We consider the vector space $\mathcal{D}=\mathbb{R}^d$. For the metric Euclidean distance, the weighted mean $(1-w)x+wy$ exists for all $x,y \ (x \not= y)$ (which is stronger than required by Lemma \ref{lemma:2}). 
%related GM can be iteratively computed by the Weiszfeld algorithm \cite{Beck2015}. 
The squared Euclidean distance $\delta^2(p,q) = ||p-q||^2$ is no longer a metric.
}}

\example{\additional{Given the weighted mean discussed above for strings, graphs, clusterings, and graph correspondences, the related metric distance function powered with integer $p \geq 2$ is no longer a metric.
}}

\vspace{2mm}
\revised[1.12]{The two lemmas above highlight the consequences of certain local properties, such as the existence of a single triple fulfilling the weighted mean requirement in the second case. An adaptation of a metric distance function can lead to a non-metric one, particularly when using the power function.}

\revised[1.11]{Lemma \ref{lemma:1} shows that a non-metric distance can still remain robust for GM computation. To provoke non-robustness, a stronger assumption about the weighted mean is required.}

\begin{theorem}
We consider a continuous space $\mathcal{D}$ with associated metric distance $\delta$. It is assumed that for any $x$, $z$ $(x \not=z)$ there exists a weighted mean $y$ for all $0 \leq w \leq 1$, Then, the GM for the non-metric distance $\delta^p$, $p$ integer $\ge 2$, is not robust and has a breakdown point \revised[2.15c]{$\epsilon^* = \frac{1}{n}$ with $\displaystyle\lim_{n\rightarrow \infty} \epsilon^* = 0$}.
\label{theorem:non-metric}
\end{theorem}

\noindent\additional{\emph{Proof sketch.} The proof is shown by a simple example of $n$ objects with one object $x$ repeating $n-1$ times and another corrupted object $y$. In addition, the proof is extended to the general case of $n-1$ inliers and one outlier $y$. In both cases the single corrupted object $y$ suffices to produce infinitely
large error of GM.}
\proofendspace

\example{\additional{We consider real numbers ($\mathcal{D}=\mathbb{R}$) with the metric distance function $\delta(p,q) = |p-q|$. The weighted mean is simply computed by $y=(1-w)x+wz$. For the non-metric distance function $\delta(p,q) = (p-q)^2$, the GM is the arithmetic mean and thus not robust. For instance, a particular number $x$ repeating $n-1$ times and another corrupted number $y$ lead to the WM: $\frac{(n-1)x+y}{n}$, which is $\frac{|x-y|}{n}$ apart from $x$. Thus, a single corrupted number $y$ suffices to produce infinitely large error of GM.}}

\additional{When considering the vector space $\mathcal{D}=\mathbb{R}^d$ and the metric Euclidean distance, the related GM can be iteratively computed by the Weiszfeld algorithm \cite{Beck2015}. The squared Euclidean distance is non-metric, resulting in the non-robust average vector. \qed}

\vspace{2mm}
Theorem \ref{theorem:non-metric} is formulated for continuous spaces where a weighted mean exists for any weighting factor $0 \leq w \leq 1$. However, the situation differs slightly in discrete spaces. In such cases, the weighted mean typically exists for a discrete but dense set of weighting factors, $w = i/K \in [0, 1], \ i=0, 1, \ldots, K$, where the path from $x$ to $y$ is sampled into $(K+1)$ objects.

\begin{corollary}
\label{corollary_discrete}
We consider a discrete space  $\mathcal{D}$ with associated metric distance $\delta$ and assume that for any $x$, $z$ $(x \not=z)$ there exists a weighted mean $y$ for densely sampled $w \in [0, 1]$, Then, the GM for the non-metric $\delta^p$, $p$ integer $\ge 2$, is not robust and has a breakdown point \revised[2.15c]{$\epsilon^* = \frac{1}{n}$ with $\displaystyle\lim_{n\rightarrow \infty} \epsilon^* = 0$}.
\end{corollary}

\noindent\additional{\emph{Proof sketch.} The proof is done by detailed consideration of the discrete case.}
\proofendspace

\example{\additional{We consider the set of all integers with the metric distance function $\delta(p,q) = |p-q|$. In this case, the WM remains the median, as it does for real numbers. However, for the squared non-metric distance function, the WM becomes the rounded arithmetic mean, which is not robust.}}

%%%%%%

\section{Proof of Theorems}
\label{sec:proofs}

\subsection{Proof of Theorem \ref{theorem:riemann_bound}}

\proof \revised[1.6, 1.17]{This proof is an adaption of the proof for the breakdown point of $0.5$ in the case of Riemannian manifolds given in \cite{fletcher2009geometric}, which is a direct generalization of the proof in \cite{lopuhaa_breakdown_1991} for Euclidean spaces.}

% definitions
The proof relies on some definitions as follows. Given a corrupted set $Q = \{q_1, ..., q_n\}$ as shown in Eq.~(\ref{eq:breakdown}), where $k < \lfloor (n-1)/2 \rfloor$ objects from an original set $O$ were replaced by random outliers, we define some auxiliary variables as shown in Figure \ref{fig:inequality_riemann}. These variables are: $\bar{q}$ is the GM of $Q$, $R = \max_i \delta(o_i,\bar{o})$ is the maximum distance of any object $o_i \in O$ to the median of $O$, $B = \{a \in \mathcal{D} : \delta(a,\bar{o}) \leq 2R\}$ is a ball of radius $2R$ around $\bar{o}$.

% argument used in the original proofs
It is then argued that for $\gamma = \inf_{a \in B} \delta(a, \bar{q})$, which is the distance of $\bar{q}$ to ball $B$,
\begin{align}
    \delta(\bar{o}, \bar{q}) &\leq 2R + \gamma \label{eq:riemannian_inequality1}\\
    \delta(\bar{q}, o_i) &\geq \delta(o_i, a) + \delta(a, \bar{q}) \nonumber \\
    &\geq R + \gamma \label{eq:riemannian_inequality2}
\end{align}
must hold. The final proof lies in showing that $\gamma$ is bounded.

% difference between proofs
\revised[1.6, 1.17, 1.19, 1.20]{In the original work by Lopuhaä et al. \cite{lopuhaa_breakdown_1991}, these relationships were shown for the Euclidean space with the metric $L_1$ norm only. As such, $\delta(o_a, o_b) = \lVert o_a - o_b \rVert$. Furthermore, the translation invariance of the median in Euclidean space was used to fix the median $\bar{o}$ as well as $B$ to the origin for a simplified proof.
This was extended by Fletcher et al. \cite{fletcher2009geometric} who noted that the proof does not use any Euclidean space specific properties and can be relaxed to metrics in Riemannian spaces easily. The authors replaced the $L_1$-norm with an arbitrary metric $\delta(\bar{o}, \bar{q})$ and removed the need for $\bar{o}$ and $B$ to be at the origin. The remainder of the proof follows the exact same structure of the original proof as only metric properties and the completeness of the space were used.
As we deal with arbitrary spaces, however, we need to adapt the proof further. In the general case, is not guaranteed that $\gamma=\inf_{a \in B} \delta(a, \bar{q})$ is the distance to the border of ball $B$ and instead realized by a point $a$ strictly inside of ball $B$. Then, the inequality (\ref{eq:riemannian_inequality2}) breaks down as illustrated in Figure \ref{fig:example_discrete}. In this case, it can be that $\delta(\bar{q}, o_i) < R + \gamma$.}

\begin{figure}[t]
    \centering
    \begin{tikzpicture}[%
    xscale=1,
    yscale=1,
    dot/.style={circle, minimum size=2pt,inner sep=0mm,draw=gray,fill=gray},
    setO/.style={circle,minimum size=5pt,inner sep=0mm,draw=black,fill=black},
    setP/.style={circle,minimum size=5pt,inner sep=0mm,draw=black},
    setExample/.style={circle, minimum size=3pt, inner sep=0pt, draw=black, fill=black}]

% grid
\foreach \x in {-2,...,5} {
    \foreach \y in {-2,...,-1} % lower half
        \node[dot] () at (\x, \y) {};
    \foreach \y in {1,...,2} % upper half
        \node[dot] () at (\x, \y) {};
}
\foreach \x in {-2, -1, 0, 1, ,2, 4, 5}  % row with missing point
    \node[dot] () at (\x, 0) {};

% X
\node at (0,1.7) {$O$};
\node[setO] (o1) at (1,1) {};
\node[setO] (o2) at (1,-1) {};
\node[setO] (o3) at (-1,1) {};
\node[setO] (o4) at (-1,-1) {};
\node[setO] (o4) at (1,0) {};
\node[setO] (o4) at (-1,0) {};

\node[setO,label=left:$\bar{o}$] (xm) at (0,0) {};

\draw[fill=black,opacity=0.1] (-1.5,-1.5) rectangle (1.5,1.9);

\begin{scope}
    \clip (-1.5,-2.2) rectangle (3,2.2);
    \draw (xm) circle[radius=1.4];
    \draw (xm) circle[radius=2.8];
\end{scope}

%\draw[dashed] (xm) -- node[below left] {$R$} (o2);
%\draw[dashed] (xm) -- node[above,pos=0.5] {\footnotesize $R$} +(1.25,0);
%\draw[dashed] (xm) -- node[above,pos=0.75] {\footnotesize $R$} +(2.5,0);

% Q
\node[setP,label=right:$\bar{q}$] (qm) at (4,0) {};

% a
\node[setExample, label=below:$a$] (a) at (2,0) {};

% R
\draw[dashed] (1,0) -- node[below] {$< R$} (2,0);

% gamma
\draw[solid] (2,0) --node [below] {$\gamma$} (qm);

% background
\draw[fill=black,opacity=0.1] (-2.2,-2.2) rectangle (5.2,2.2);

% R
%\draw[dashed] (qm) -- node[above] {\footnotesize $\gamma$} +(-1.5,0);

% a inside
%\draw (qm) -- node[below] {$\gamma$} (a);
%\draw[] (a) -- 
%    node[below, pos=0.75] {\footnotesize $R$} 
%    node[below, pos=0.3] {\footnotesize $<R$} 
%    node[setExample, pos=0.47] {} 
%    (xm);

\end{tikzpicture}
    \caption{\revised[1.6, 1.17, 1.19, 1.20]{Proof of Theorem \ref{theorem:riemann_bound}. Illustration of the breakdown of Inequality (\ref{eq:riemannian_inequality2}) in the case of discrete spaces. In the above example, objects only lie on a regular grid. In this case, point $a$ in the definition of $\gamma$ is strictly inside of ball $B$. This can lead to the distance $\delta(\bar{q}, o_i)$ being shorter than $R+\gamma$.}}
    \label{fig:example_discrete}
\end{figure}

% changes needed for arbitrary discrete spaces
\revised[1.6, 1.17, 1.19, 1.20]{
Therefore, instead assume that two points $a$ and $b$ as shown in Figure \ref{fig:inequality_riemann} exist with the following properties:
\begin{align*}
    \delta(\bar{o},a) &\leq 2R \leq \delta(\bar{o},b)\\
    \delta(a,b) &\leq c
\end{align*}
with $0 \leq c \leq \infty$ being a constant. In other words, point $a$ lies inside ball $B$ with radius $2R$, point $b$ lies outside ball $B$, and the distance between $a$ and $b$ is smaller than a constant $c$. Then, we define $\gamma' = \delta(b, \bar{q})$ as the distance of $b$ to the corrupted median. In essence, points $a$ and $b$ are bridging the gap between points on the inside and outside of ball $B$ by at most $c$.}

Then, by triangle inequality we obtain for (\ref{eq:riemannian_inequality1})
\begin{align}
    \delta(\bar{o}, \bar{q}) &\leq \delta(\bar{o}, a) + \delta(a,b) + \delta(b,\bar{q}) \nonumber \\
    &\leq 2R + c + \gamma' \label{eq:discrete_inequality1}
\end{align}
while for (\ref{eq:riemannian_inequality2}), it remains
\begin{align}
    \delta(\bar{q}, o_i) &\geq \delta(o_i, a) + \delta(a, b) + \delta(b, \bar{q}) \nonumber \\
    &\geq R + c + \gamma' \nonumber \\
    &\geq R + \gamma' \nonumber \\
    &\geq \delta(\bar{o},o_i) + \gamma' \label{eq:discrete_inequality2}
\end{align}
Using a proof by contradiction, we can show that $\gamma'$ cannot grow indefinitely, which leads to the breakdown point. In the following we will assume the worst case in the displacement of $\bar{q}$, especially a displacement of $\bar{q}$ outside of ball $B$. In the case that $\bar{q}$ lies inside of $B$, the displacement is finite and Theorem \ref{theorem:riemann_bound} would be true.

Using the triangle inequality again, one can additionally see:
\begin{align*}
 \delta(\bar{q}, q_i) &\geq \delta(\bar{o},q_i) - \delta(\bar{o},\bar{q}) \geq \delta(\bar{o},q_i) - (2R+c+\gamma')
\end{align*}
Combining above with (\ref{eq:discrete_inequality2}) and the fact that $n-k = n - \lfloor(n-1)/2\rfloor$ of the $q_i$ are from the original set $O$ and $\lfloor(n-1)/2\rfloor$ are not from $O$ leads to:
\begin{align*}
    \sum_{i=1}^n \delta(\bar{q}, q_i) \geq& \sum_{i=1}^n \delta(\bar{o}, q_i) - \underbrace{\lfloor (n-1) / 2\rfloor(2R + c + \gamma')}_{q_i \notin O} \\
    & + \underbrace{(n-\lfloor(n-1)/2\rfloor)\gamma'}_{q_i \in O}\\
    \geq& \sum_{i=1}^n \delta(\bar{o},q_i) - \lfloor (n-1)/2\rfloor (2R+c) + \gamma' \\
\end{align*}

\vspace{-7mm}
At this point, assuming $\gamma' > \lfloor(n-1)/2\rfloor (2R+c)$ leads to:
\begin{align*}
    \sum_{i=1}^n \delta(\bar{q},q_i) &> \sum_{i=1}^n \delta(\bar{o},q_i)
\end{align*}
This is a contradiction to the fact that $\bar{q}$ is the median of $q_i$. Therefore, $\gamma' \leq \lfloor (n-1)/2\rfloor (2R+c)$ and the maximum displacement is bounded by:
\begin{align}
    \delta(\bar{q},\bar{o}) &\leq 2R + c + \gamma' \ \leq \lfloor (n+1)/2 \rfloor (2R+c) \nonumber\\
    &\leq (n+1)R + \lfloor(n+1)2\rfloor c \label{eq:riemann_bound} \\
    &< \infty \nonumber
\end{align}
for all $k \leq \lfloor(n-1)/2 \rfloor$.
The breakdown point is thus $\epsilon^* \geq \lfloor (n+1)/2 \rfloor/n$. Note this proof only requires $c$ being a constant, the existence of any points inside and outside of radius $2R$ is sufficient for this bound to be finite. A small $c$ is only relevant for finding an upper bound of the displacement itself through (\ref{eq:riemann_bound}).

It can be easily shown as follows that the case $\epsilon^* = \lfloor (n+1)/2 \rfloor/n$ can be reached and is therefore a tight lower bound. Given a set $O \subseteq \mathbb{R}$, corrupt $k \geq \lfloor(n+1)/2 \rfloor$ objects by moving them to position $d \in \mathbb{R}$. Then, the GM will be $\bar{q} = d$. Since $d$ can be chosen arbitrarily large, $\inf_d \delta(\bar{o},\bar{q}) = \infty$ and $k$ therefore cannot be the solution to $\epsilon^*$.
Hence,
\begin{equation*}
    \begin{gathered} % for multiline centered equation
    \epsilon^* = \lfloor (n+1)/2 \rfloor/n \\
    \lim_{n \rightarrow \infty} \epsilon^* = \lim_{n\rightarrow\infty} \lfloor (n+1)/2\rfloor/n = 0.5
    \end{gathered}
\end{equation*}
in $\mathbb{R}$ using the Euclidean distance.  \qed
%\end{IEEEproof} \proofendspace

\subsection{Proof of Theorem \ref{theorem:added_bound}}
\proof
Due to the definition of the median as the minimizer of the sum of distance $\SOD$ in Eq.~ (\ref{eq:median}), it holds:
\begin{align}
	\SOD_O(\bar{o}) &\leq \SOD_O(\bar{q}) \label{eq:1}\\
%\intertext{and}
	\SOD_Q(\bar{q}) &\leq \SOD_Q(\bar{o}) \label{eq:2}\\
	\Leftrightarrow \SOD_O(\bar{q}) + \SOD_P(\bar{q}) &\leq \SOD_O(\bar{o}) + \SOD_P(\bar{o}) \label{eq:3}\\
\intertext{Combining Eqs. (\ref{eq:1}) and (\ref{eq:3}) leads to:}
	\SOD_P(\bar{q}) &\leq \SOD_P(\bar{o}) \label{eq:4}
\end{align}
Using this we can derive, starting with Eq.~(\ref{eq:2}):
\begin{align}
	\SOD_Q(\bar{q}) &\leq \SOD_Q(\bar{o}) \nonumber\\
	\Leftrightarrow \SOD_O(\bar{q}) &\leq \SOD_O(\bar{o}) + \SOD_P(\bar{o}) - \SOD_P(\bar{q}) \nonumber\\
	& = \SOD_O(\bar{o}) + \sum_{i=1}^{k} (\delta(\bar{o},p_i) - \delta(p_i,\bar{q})) \label{eq:5}\\
\intertext{Due to Eq.~(\ref{eq:4}), the sum in Eq.~(\ref{eq:5}) is always positive. Therefore, we can apply the reverse triangle inequality to each summand:}
	\SOD_O(\bar{q}) &\leq \SOD_O(\bar{o}) + k \cdot \delta(\bar{o},\bar{q}) \label{eq:6}
\end{align}
This relationship can be seen in Figure \ref{fig:inequality_added}. $\SOD_O(\bar{q})$ (dotted lines) is bounded by $\SOD_O(\bar{o})$ (solid lines) and $k \cdot \delta(\bar{o},\bar{q})$ (dashed line).
On the other hand, using the triangle inequality $n$ times on $\delta(\bar{q},\bar{o})$ leads to:
\begin{align}
	n \cdot \delta(\bar{q},\bar{o}) &\leq \sum_{i=1}^{n} \delta(\bar{q},o_i) + \sum_{i=1}^{n} \delta(o_i,\bar{o}) \nonumber\\
    & = \SOD_O(\bar{q}) + \SOD_O(\bar{o}), \nonumber\\
\intertext{again illustrated in Figure \ref{fig:inequality_added}. $n \cdot \delta(\bar{q},\bar{o})$ (dashed line) is bounded by $\SOD_O(\bar{q})$ (dotted lines) and $\SOD_O(\bar{o})$ (solid lines). Inserting Eq.~(\ref{eq:6}) under the assumption of $k < n$ leads to}
	n \cdot \delta(\bar{q},\bar{o}) &\leq 2\;\SOD_O(\bar{o}) + k \cdot \delta(\bar{o},\bar{q}) \nonumber\\
	\Leftrightarrow (n-k) \cdot \delta(\bar{o},\bar{q}) &\leq 2\;\SOD_O(\bar{o})\nonumber\\
	\Leftrightarrow \delta(\bar{o},\bar{q}) &\leq \frac{2}{n-k} \SOD_O(\bar{o}) \label{eq:7}
\end{align}
and Theorem \ref{theorem:added_bound} is proven. \qed

\subsection{Proof for Theorem \ref{theorem:replaced_bound}}
\proof
Let $A = O \setminus Q$ and $B = Q \setminus O$ be the set of replaced objects and corrupted objects with $|A| = |B| = k$, respectively. Since $|X| = n-k > k$, $O = X \bigcup A$ and $Q = X \bigcup B$, Theorem \ref{theorem:added_bound} leads to:
\begin{align*}
    \delta(\bar{o},\bar{q}) &\leq \delta(\bar{x},\bar{o}) + \delta(\bar{x},\bar{q})\\
    &\leq \frac{2}{(n-k)-k} \SOD_X(\bar{x}) + \frac{2}{(n-k)-k} \SOD_X(\bar{x})\\
    &= \frac{4}{n-2k} \SOD_X(\bar{x}) \label{} \qed
\end{align*}

\subsection{Proof for Theorem \ref{theorem:sod_bound}}
The Proof is a direct application of the triangle inequality and Theorem \ref{theorem:added_bound}.
\proof
\revised[2.12]{Since $Q = O \bigcup P$}, and $\SOD_O(\bar{o}) < \SOD_O(\bar{q})$, it follows:
\begin{align*}
    & \SOD_Q(\bar{o}) - \SOD_Q(\bar{q})  \ \leq \ \SOD_P(\bar{o}) - \SOD_P(\bar{q})\\
    &= \sum_{p_i \in P} \delta(p_i,\bar{o}) - \sum_{p_i \in P} \delta(p_i,\bar{q})
    \ \leq \ \sum_{p_i \in P} \delta(\bar{o},\bar{q})\\
    &\leq \sum_{p_i \in P} \frac{2}{n-k} \SOD_O(\bar{o}) \mbox{ (because of Theorem \ref{theorem:added_bound})}\\
    &=\frac{2k}{n-k} \SOD_O(\bar{o}) \qed
\end{align*}

\subsection{Proof for Theorem \ref{theorem:added_bound_weighted}}
\proof
The proof follows the proof of Theorem \ref{theorem:added_bound}. As the GM minimizes $\SOD$, Eqs. (\ref{eq:1}) to (\ref{eq:4}) hold for the weighted case as well. From this we can derive (compare Eq. (\ref{eq:5}))
\begin{align}
    \SOD_O^W(\bar{q}) &\leq \SOD_O^W(\bar{o}) + \SOD_P^W(\bar{o}) - \SOD_P^W(\bar{q}) \nonumber \\ 
     & = \SOD_O^W(\bar{o}) + \sum_{p_i \in P} w_{p_i} (\delta(\bar{o},p_i) - \delta(p_i,\bar{q})) \nonumber \\ 
     &\leq \SOD_O^W(\bar{o}) + \sum_{p_i \in P} w_{p_i} \delta(\bar{o}, \bar{q}) \label{eq:wm1}
\end{align}
On the other hand:
\begin{align}
    \sum_{o_i \in O} w_{o_i} \delta(\bar{q},\bar{o}) &\leq  \sum_{o_i \in O} w_{o_i} \delta(\bar{q}, o_i) + \sum_{o_i \in O} w_{o_i} \delta(o_i, \bar{o}) \nonumber\\
    & = \SOD_O^W(\bar{q}) + \SOD_O^W(\bar{o}) \nonumber
\end{align}
Inserting \revised[2.13c]{Eq. (\ref{eq:wm1})} under the assumption of $\sum_{p_i \in P} w_{p_i} < \sum_{o_i \in O} w_{o_i}$ leads to
\begin{align*}
    \sum_{o_i \in O} w_{o_i} \delta(\bar{q},\bar{o}) &\leq 2 \SOD_O^W(\bar{o}) + \sum_{p_i \in P} w_{p_i} \delta(\bar{o}, \bar{q}) \\
    \Leftrightarrow \delta(\bar{o},\bar{q}) &\leq \frac{2}{\sum_{o_i \in O} w_{o_i} - \sum_{p_i \in P} w_{p_i}} \SOD_O^W(\bar{o})
\end{align*}
and Theorem \ref{theorem:added_bound_weighted} is proven. \qed

\proofendspace

\subsection{Proof for Theorem \ref{theorem:replaced_bound_weighted}}
\proof
Let $A = O \setminus Q$ be the set of replaced objects. Since the weights are not changed, $\sum_{a_i \in A} w_{a_i} = \sum_{p_i \in P} w_{p_i} < \sum_{x_i \in X} w_{x_i}$. Therefore, Theorem \ref{theorem:added_bound_weighted} leads to:
\begin{align*}
    \delta(\bar{o},\bar{q}) &\leq \delta(\bar{x},\bar{o}) + \delta(\bar{x},\bar{q})\\
    &\leq \frac{2}{\sum_{x_i \in X} w_{x_i} - \sum_{a_i \in A} w_{a_i}} \SOD_X(\bar{x}) +\\ &\phantom{=}\frac{2}{\sum_{x_i \in X} w_{x_i} - \sum_{p_i \in P} w_{p_i}} \SOD_X(\bar{x})\\
    &\leq \frac{4}{\sum_{x_i \in X} w_{x_i} - \sum_{p_i \in P} w_{p_i}} \SOD_X(\bar{x}) \qed
\end{align*}

\subsection{Proof for Theorem \ref{theorem:non-metric}}
\proof
\additional{We demonstrate the non-robustness for two cases, both with $n$ objects:
\begin{itemize}
\item \textbf{Simplified case}: one object $x$ repeating $n-1$ times and one corrupted object $y$, see Figure \ref{fig:non_metric}a).
\item \textbf{General case}: $n-1$ inlier objects $x_1, \ x_2, \ldots, x_{n-1}$ and one corrupted object $y$, see Figure \ref{fig:non_metric}b).
\end{itemize}
}

\noindent\additional{\textbf{Simplified case}}:
The proof consists of two parts. We first show that in this case the GM must be a weighted mean of $x$ and $y$. This result is then used to demonstrate the non-robustness of this particular setting of $n$ objects.

\vspace{2mm}
\noindent{\textbf{Part 1:}}
We show the first result by contradiction. The GM is assumed to be $H$ with $\delta(x,H)=A$ and $\delta(H,y)=B$. Since $\delta$ is a metric, it holds:
\begin{equation}
\label{eqn_1}
A + B \ > \ \delta(x,y) = d
\end{equation}
Note that $A+B=d$ is excluded due to the assumption that $H$ is not a weighted mean of $x$ and $y$, thus $H\not=y$ holds.
\revised[2.15(d)]{There are two possible situations ($A \geq d$, $A < d$), in each of which it holds $\Omega(h) \geq \Omega(H)$ for all weighted mean $h$ of $x$ and $y$. However, we show that there exists a counterexample in each case.}
\begin{itemize}
\item $A \geq d$. We study the particular weighted mean $h=y$ and compute the related sum of distance.
\begin{eqnarray*}
\Omega(h) & = & (n-1) a^p + b^p \ = \ (n-1) d^p \\
& < & (n-1) A^p + B^p \ [B>0 \makebox{ because } H\not=y] \\
& = & \Omega(H)
\end{eqnarray*}
\item $A < d$. We study the particular weighted mean $h$ such that $\delta(x,h)=A$ and compute the related sum of distance.
\begin{eqnarray*}
\Omega(h) & = & (n-1) a^p + b^p \ = \ (n-1) A^p + (d-A)^p\\
& < & (n-1) A^p + B^p \ [\makebox{because } (\ref{eqn_1})] \\
& = & \Omega(H)
\end{eqnarray*}
\end{itemize}
In both situations the derived result $\Omega(h) < \Omega(H)$ is a contradiction to the assumption that $H$ is a GM. Therefore, the GM must be a weighted mean of $x$ and $y$. 

\vspace{2mm}
\noindent{\textbf{Part 2:}}
Given the last result we now \revised[2.15e]{figure out the weighted mean $h$ of $x$ and $y$ that corresponds to the GM, which must minimize $\Omega(h)$. This is done} by setting the derivative of the related sum of distance $\Omega(h)$ to zero:
\begin{eqnarray*}
\frac{d\Omega(h)}{da} & = & \frac{d}{da} \ [(n-1)a^p + (d-a)^p]
 \\
 & = & p(n-1)a^{p-1} -p(d-a)^{p-1} \\
 & = & 0
 \end{eqnarray*}
\revised[2.15e]{Here the differentiability of the distance function $\delta$ is assumed}. This equation has the easy-to-derive solution:
 \begin{equation}
 a^* \ = \ \frac{d}{(n-1)^{1/(p-1)}+1}
 \label{eqn_a}
\end{equation}
\revised[2.15e]{In addition, the derivative $\frac{d\Omega(h)}{da}$ is negative for $a<a^*$ and positive for $a>a^*$. Thus, we obtain the global miminmum with $a=a^*$.}
A single corrupted object $y$ suffices to produce infinitely large error of GM, resulting in the breakdown point \revised[2.15c]{$\epsilon^* = \frac{1}{n}$ with $\displaystyle\lim_{n\rightarrow \infty} \epsilon^* = 0$}.

\noindent\additional{\textbf{General case}:
We assume the GM of the $n-1$ inliers to be $\bar{x}$. These objects are bounded within a maximal distance $r$ relative to $\bar{x}$. Since the breakdown focuses on the behavior of outliers that are infinitely far from $\bar{x}$, we define a new distance function:
\[
\delta^*(x,y) \ = \ \frac{\delta(x,y)}{r} \cdot \epsilon
\]
where $\epsilon$ is a small constant (close to zero). $\delta^*(x,y)$ is merely a scaled version of $\delta(x,y)$. Thus, the GM remains unchanged. Intuitively, this scaling simulates a "zooming out" effect, such that the $n-1$ inliers are now bounded by $\epsilon$ only from $\bar{x}$. We can then approximate the inliers as if they all coincide at $\bar{x}$ (i.e. $\epsilon \rightarrow 0$), which exactly corresponds to the simplified case. Consequently, a single corrupted object $y$ is sufficient to produce an infinitely large error of GM, leading to a zero breakdown point for $n\rightarrow \infty$. Note that the approximation used above is only possible due to the infinite nature of breakdown point.
}
\qed

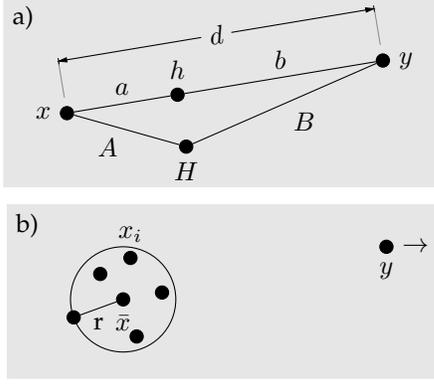
\begin{figure}
    \centering
    % needs 
% \usepackage{tikz-dimline}
% \usetikzlibrary{calc}

\begin{tikzpicture}[%
    point/.style={circle,minimum size=5pt,inner sep=0mm,draw=black,fill=black},
    xscale=0.7,
    yscale=0.7]
    
% label
\node[anchor=north west] () at (-2.2, 2.2) {a)};

% Points
%\node[point,label={[label distance=-5mm]200:\shortstack{$x$\\{\small $(n-1)$ times}}}] (x) at (0,0) {};
\node[point,label=left:$x$] (x) at (-1,0) {};
\node[point,label=right:$y$] (y) at (5,1) {};
\node[point,label=above:$h$] (h) at ($(x)!0.35!(y)$) {}; 
% ratio 0.35 from (x) to (y)

\node[point,label=below:$H$] (H) at ($(h)!10mm!-90:(y)$) {}; 
% 15mm distance from (h) to the direction of (y), direction rotated by -90 degree

% lines between points
\draw (x) -- node[above]{$a$} (h);
\draw (h) -- node[above]{$b$} (y);
\draw (x) -- node[below left]{$A$} (H);
\draw (y) -- node[below right]{$B$} (H);

% distance line
\dimline[extension start length=7mm, extension end length=7mm]{($(x)!10mm!90:(y)$)}{($(y)!10mm!-90:(x)$)}{$d$};

% Background
\draw[fill=black,opacity=0.1] (-2.2,-1.4) rectangle (6,2.2);
%\draw[fill=black,opacity=0.1] (-2.2,-2.2) rectangle (6,2.2);

\end{tikzpicture}
    \vspace{2mm}
    % \usetikzlibrary{calc}

\begin{tikzpicture}[%
    point/.style={circle,minimum size=5pt,inner sep=0mm,draw=black,fill=black},
    xscale=0.7,
    yscale=0.7]

    % label
    \node[anchor=north west] () at (-2.2, 1.8) {b)};

    % median
    \node[point, label=below:$\bar{x}$] (xm) at (0,0) {};

    % set_x: (angle:distance) notation
    \node[point, label=above:{$x_i$}] () at ($(xm)+(80:0.8)$) {};
    \node[point] () at ($(xm)+(132:0.65)$) {};
    \node[point] () at ($(xm)+(290:0.75)$) {};
    \node[point] () at ($(xm)+(10:0.75)$) {};

    % border pointa
    \node[point] (rp) at ($(xm)+(200:1)$) {};

    % circle and radius
    \draw[] (xm) circle (1);
    \draw[] (xm) to node[below] {r} (rp);

    % y
    \node[point, label=below:$y$, label=right:$\rightarrow$] (y) at (5,1) {};

    % background
    \draw[fill=black,opacity=0.1] (-2.2,-1.5) rectangle (6,1.8);
\end{tikzpicture}
    \caption{Proof of Theorem \ref{theorem:non-metric}. Although drawn in a plane, this illustration should be understood as a general space. \additional{a) Simplified case: $x$ is repeated $n-1$ times. b) General case: Although $x_i$ are in a radius $r$ around GM $\bar{x}$, the deviation is insignificant against the infinitely far away $y$ as symbolized by the arrow.}}
    \label{fig:non_metric}
\end{figure}

\subsection{Proof for Corollary \ref{corollary_discrete}}
\proof
Part 1 of the proof for Theorem \ref{theorem:non-metric} applies to the discrete case as well. \revised[2.15f,g]{For part 2 the solution differs slightly. Here the weighted mean of $x$ and $y$ does not build a continuum, but instead a discrete set $WM$, each member of which has a related weighting factor. Generally, the optimal weighting factor $a^*$ as computed by (\ref{eqn_a}) is not a member of this set. Let $w_l, \ w_r$ be the weighting factor of the weighted mean directly left and right to $a^*$, respectively. Due to the monotony of the derivative $\frac{d\Omega(h)}{da}$ on each side of $a^*$, respectively, the GM must correspond to one of them that has smaller $\Omega()$. We assume that the discrete set $WM$ is sufficiently dense such that the GM corresponds to the weighting factor:
\begin{equation*}
 a^* \ = \ \frac{d}{(n-1)^{1/(p-1)}+1} + \epsilon
\end{equation*}
where $\epsilon$ is a small term relative to $d$}. Given this fact, a single corrupted object $y$ suffices to produce infinitely large error of GM. The breakdown point is thus \revised[2.15c]{$\epsilon^* = \frac{1}{n}$ with $\displaystyle\lim_{n\rightarrow \infty} \epsilon^* = 0$ in the discrete case as well.} \qed
%\end{IEEEproof}

%%%%%%
\section{Applicability of theoretical results}
\label{sec:application}

\additional{This section first addresses general considerations regarding the applicability of our results. We then present two experimental validations: the averaging of 3D rotations and the averaging of rankings.}

\subsection{General considerations}

In addition to theoretical interest the results shown in the paper also have general impact and applicability. Here we briefly discuss some application scenarios.

A breakdown point $\geq 0.5$ means that even if 50\%  or more of the objects in the input set are outliers or otherwise corrupted, the GM computation still leads to a result that is relatively close (i.e. does not diverge arbitrarily) to the consensus object of the original set without outliers. This fact increases the certainty in dealing with typically unavoidable outliers in many practical situations.

In Section \ref{sec:weighted-med} (robustness of weighted GM) we assumed the worst case of largest weights belonging to the outliers. However, in practice this is often not the case. Higher weights are usually assigned to more certain results (e.g. classifiers with a higher accuracy), meaning the likelihood of an object being an outlier decreases with increasing weights. This directly leads to a weighted GM that is often more robust than our assumed worst case. Our theoretical results thus provide deeper understanding of weighted GM computation and support the design of corresponding computation algorithms.

Our work also helps combat the lacking awareness of non-robust behavior of non-metric distance functions. The special class of non-metric distance functions $\delta^p$ ($p$ integer $\ge 2$) with non-robust behavior studied in Section \ref{sec:non-metric} appeared, in particular $p=2$, again and again in the literature, often with no particular justification (e.g. \cite{mcmorris2012}). Typically, it was experimentally observed that such computation with $\delta^p$ is less robust than using a metric distance function $\delta$ \cite{Dai2009,Cunha_2019,Mankovich2022} (e.g. "we discover that the flag median [$p=1$] is the most robust to
outliers and produces higher cluster purities than the flag mean [$p=2$]" \cite{Mankovich2022}).
Our results help avoid such non-robust computation, without the need of experimental work to show the non-robustness as done before.
%\textcolor{red}{\cite{tang2021} - lp-estimator: minimizing SOD $\delta^p$ with specific $\delta$}
%\textcolor{red}{\cite{mcmorris2012} - lp-estimator: different properties of median with $\delta^p$. $\delta$ is the shortest path distance. focuses on $p \geq 2$}

\additional{On the other hand, there are situations where using a non-robust distance function like $\delta^p$ is practically harmless, such as in atlas construction for medical imaging. One example is \cite{atlas_2013}, where the atlas construction is formulated as GM computation. In these cases, the setting is controlled: the data are typically checked beforehand to ensure the atlas is reasonable, and as a result, there are no outliers in the data. Even in such cases, our results are useful. They help users recognize the potential risks and understand why it may still be acceptable to use a non-robust distance function.}

\subsection{\revised[1.23, 2.1, 2.6, 2.16]{Robust averaging of 3D rotations}}

\begin{figure}
    \centering
    \hspace*{-3mm}
    \includegraphics[width=1\linewidth]{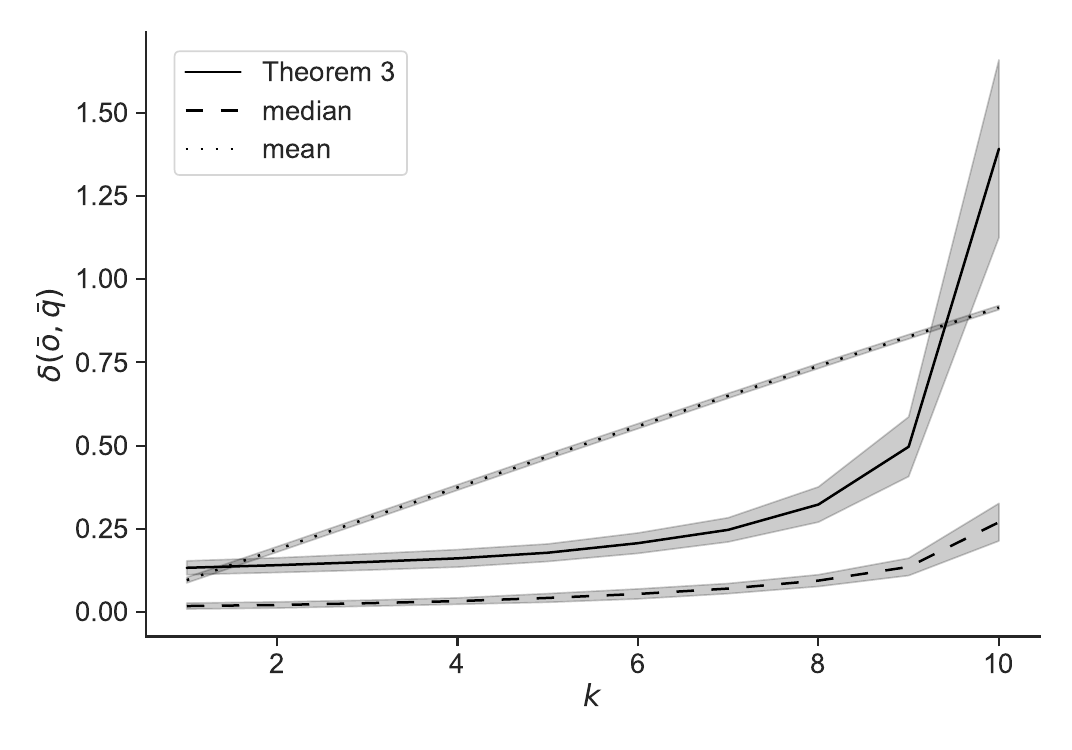}
    \vspace{-1em}
    \caption{\revised[2.1, 2.6, 2.16]{Displacement of the rotation mean and median $\delta(\bar{o}, \bar{q})$ depending on $k$ outliers in a set of $n=21$ normally distributed rotations. The comparison shows the maximum displacement of the median calculated with Theorem \ref{theorem:replaced_bound} (solid) as well as the displacements of the computed median (dashed) and the computed mean (dotted).}}
    \label{fig:3d-rotations}
\end{figure}

\revised{As shown in the introduction, averaging 3D rotations is a common task for computer vision, in which several 3D rotations are estimated and combined to get a robust estimation of the target objects' true rotation.}

\revised{In Figure \ref{fig:3d-rotations} we show the difference in displacement $\delta(\bar{o}, \bar{q})$ where $\bar{q}$ is computed by mean and median 3D rotation using the angular distance shown in the introduction. In this experiment, we first randomly selected a base rotation matrix by selecting random rotation angles and constructing its corresponding $3\times3$ matrix. Then, we constructed $n=21$ rotation matrices by adding noise to the base rotation angles and constructing their respective rotation matrices. Finally, $k$ of these rotation matrices were replaced by outlier rotation matrices with large distances to the original matrix. From these corrupted sets, we computed the upper bound of the displacement of the GM (solid), the displacement of the median rotation (dashed) and the displacement of the mean rotation (dotted) between the uncorrupted set and the set where $k$ objects were replaced by outliers. Mean and median were computed using gradient descent with the set mean/median (i.e the object in the set with the smallest sum of distance) as starting point. This was repeated $20$ times for each $k$. The shown results are averages with standard deviation as shaded area.}

\revised{As can be seen, not only is the computed displacement $\delta(\bar{o}, \bar{q})$ in the median rotation much lower than the displacement of the mean rotation, but even the upper bound of the median displacement computed with Theorem \ref{theorem:replaced_bound} is much lower than the mean displacement for a large range of $k$. This shows that the GM is an effective method for robust rotation combination, and the bound in Theorem \ref{theorem:replaced_bound} shows values that are in range of true results, therefore making it a practical estimation of median error.}

\subsection{\revised[1.23, 2.1, 2.6, 2.16]{Robust averaging of rankings.}}

\begin{figure}
    \centering
    \hspace*{-3mm}
    \includegraphics[width=1\linewidth]{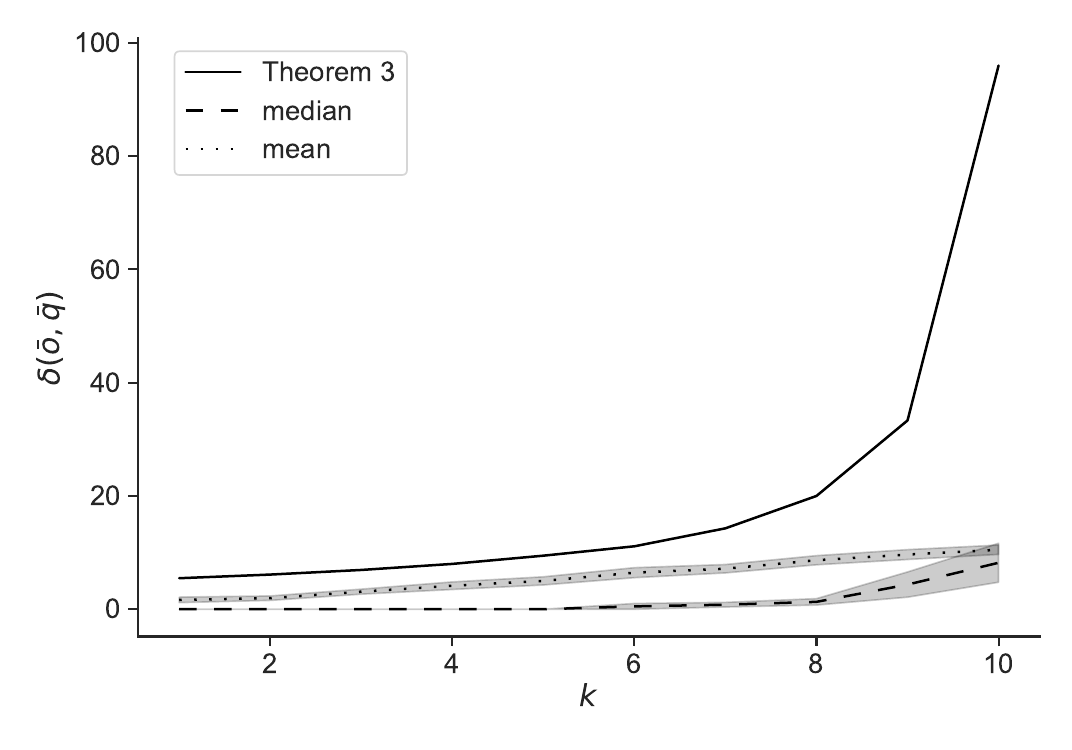}
    \vspace{-1em}
    \caption{\revised[2.1, 2.6, 2.16]{Displacement of the ranking mean and median $\delta(\bar{o}, \bar{q})$ depending on $k$ outliers of $n=21$ rankings. The comparison shows the maximum displacement of the median calculated with Theorem \ref{theorem:replaced_bound} (solid) as well as the displacements of the computed median (dashed) and the computed mean (dotted).}}
    \label{fig:rankings}
\end{figure}

\revised{As shown in the introduction, averaging of rankings is a common task in consensus learning, often using the popular Kendall-tau distance shown in the introduction.}

\revised{Figure \ref{fig:rankings} shows the difference of mean and median ranking of $n=21$ rankings of length $7$. The original ranking set was constructed by selecting a random initial ranking and adding noise by random swapping of elements. The outlier rankings were constructed similarly from a second random ranking with large distance to the first one. Then, the median and mean were computed using the Kendall-tau metric for different sets where $k$ original objects were replaced by the generated outliers. The mean was computed using the GM with $\delta^2$ as distance, while the median was computed as the GM with $\delta$. In both cases, all possible rankings of length $7$ were enumerated to find the exact mean and median. This was repeated $20$ times for each $k$. The shown results are averages with standard deviation as shaded area.}

\revised{Similar to the results for 3D rotations, the median ranking (dashed) shows a much smaller displacement from the original ranking compared to the mean ranking (dotted). For $k \leq 8$, the median even matches the original ranking exactly in most cases. The mean ranking on the other hand is not robust and linearly increases its error with growing number of outliers.
Although higher than in the 3D rotation example, the bound of Theorem \ref{theorem:replaced_bound} (solid) again shows a somewhat low maximum bound.}

%%%%%
\section{\additional{Obtaining metric distance functions}}
\label{sec:metric}

\additional{Given the importance of metric distance functions for robust GM computation we shortly discuss some techniques to obtain metrics in this section.}
\additional{One common technique for the construction of customized metrics is metric learning \cite{yang2006distance, kulis2013metric}. This technique aims to adapt existing metrics in Euclidean or Riemannian spaces to achieve specialized properties, generally that objects with similar properties (e.g. class labels) have a small distance, while objects with different properties have large distances. Thus, metric learning follows a different aim.}
%\additional{Another possibility are methods such as "Generalized Non-Metric Multidimensional Scaling" \cite{agarwal2007generalized}. This method computes a distance metric only from given orderings in the form of $\delta(o_a, o_b) < \delta(o_c, o_d)$, i.e. only the relative ordering of distances is needed.}
\additional{In the following, we summarize two cases where metrics can be constructed from other base functions.}

\subsection{\additional{Metrics from kernel functions}}

\additional{
Kernel functions \cite{Kernel_Book_2004} are positive definite symmetric functions $K\!: X \times X \rightarrow \mathbb{R}$. Kernel functions compute the scalar product of two objects $o_a$ and $o_b$ after embedding into a possibly unknown and infinite-dimensional kernel space where a scalar product is defined. They have been popular in applications such as support vector machines, where the so-called \emph{kernel-trick} allows the computation of a non-linear decision boundary function by only performing scalar product computations in this kernel space using the kernel function.}

\additional{
Given such a positive definite kernel function $K$, one can obtain a metric by:
\[
\delta(o_a,o_b) \ = \ \sqrt{K(o_a,o_a) - 2 K(o_a,o_b) + K(o_b, o_b)}
\]
i.e. by computing the norm of two objects in the kernel space. Note that kernel metric learning exists to refine this metric to adhere to specific properties \cite{yang2006distance}.}

\subsection{\additional{Metrics from arbitrary functions}}
\additional{Another possibility is to force arbitrary functions to fulfill metric properties by function transformations. A metric is required to have the following four properties: The distance from an object to itself is zero ($\delta(x,x) = 0$), positivity (for $x \neq y: \delta(x,y) > 0$), symmetry ($\delta(x,y) = \delta(y,x))$, and triangle inequality ($\delta(x,z) \leq \delta(x,y) + \delta(y,z)$).
Given an arbitrary function $f(o_a, o_b): X \times X \rightarrow \mathbb{R}$, one can ensure that these properties are fulfilled in several ways.}

\additional{Distance from an object to itself is zero: 
$$\delta(o_a, o_b) = f(o_a, o_b) - \frac{1}{2}(f(o_a,o_a) + f(o_b,o_b))$$ 
Positivity (with $c>0$): 
$$\delta(o_a, o_b) = \begin{cases}
    0 & o_a = o_b\\
    \lvert f(o_a, o_b)\rvert + c & o_a \neq o_b\\
\end{cases}$$
Symmetry:
$$\delta(o_a, o_b) = \frac{1}{2}(f(o_a,o_b) + f(o_b, o_a))$$
For the triangle inequality, there is no general transformation to ensure its fulfillment aside from trivial (but useless) ones such as:
\begin{equation*}
\delta(o_a, o_b) = \begin{cases}
    0 & f(o_a, o_b) = 0 \\
    1 & f(o_a, o_b) \neq 0
\end{cases}
\end{equation*}
where $f(o_a,o_b)$ already fulfills the above criteria. However, in special cases it is possible to ensure this property. If, for example, the space is discrete, one can use the transformation:
\begin{equation*}
\delta(x, y) = \inf \left\{ \sum_{i=1}^{n} f(o_{i-1}, o_i) \mid n \in \mathbb{N}, o_0 = o_a, o_n = o_b \right\}
\end{equation*}
to ensure the fulfillment of the triangle inequality. This is equivalent to the shortest path in a graph and therefore a metric.}

%%%%%
\section{Conclusion}
\label{sec:conclusion}

In this paper we investigated several robustness aspects of GM computation, which were open in the literature so far. We presented robustness characterization of (weighted) GM computation in a general setting, i.e. without assumption of any particular space, including the breakdown point $\geq 0.5$ of GM computation for metric distance functions in general and the detailed behavior analysis in case of outliers. In addition, we also presented some robustness results for non-metric distance functions. These results contribute to closing a gap in the research literature and have general impact and applicability, e.g. providing deeper understanding of GM computation and practical guidance to avoid non-robust computation. 

This work motivates further research on an number of open issues. In particular, the robustness characterization for non-metric distance functions is generally challenging. Our current work is limited to a particular class of non-metric distance functions only. There is thus room for extended consideration. In addition, we have shown that the computed GM is relatively close (i.e. does not diverge arbitrarily) but not equal to the consensus object without outliers. It is still necessary to develop robust algorithms that can fully cope with outliers, reducing or ideally eliminating their influence on the GM computation.
%\revised[1.15, 2.4]{Robust computation of GM is especially important when working with non-metric distance functions. One approach is drawn from robust statistics, including traditional methods such as M-estimators, least median, and least quantile \cite{Stewart1999}, as well as more recent techniques (e.g. \cite{Lecue_2020}), including their efficient solution \cite{Peng_2023}. We will particularly study robust GM computation in future.}

%%%%%%
\section*{Acknowledgments}
Andreas Nienkötter was supported by the National Natural Science Foundation of China Fund No.\ W2433165 and the Sichuan Province Key Research and Development Fund No.\ 2023YFWZ0009. Xiaoyi Jiang was partly supported by the Deutsche Forschungsgemeinschaft (DFG: CRC 1450 – 431460824, SPP2363) and European Union’s Horizon 2020 research and innovation programme under the Marie Sklodowska-Curie grant agreement No 778602 Ultracept.

% or


\ifCLASSOPTIONcaptionsoff
  \newpage
\fi



%\bibliographystyle{IEEEtran}

\bibliographystyle{IEEEtran}
\bibliography{Literature}

\end{document}